\documentclass[onecolumn]{robbyant}

\metadata[Website]{\url{https://technology.robbyant.com/lingbot-map-v2}}
\metadata[Github]{\url{https://github.com/robbyant/lingbot-map-v2}}
\metadata[Checkpoints]{\url{https://huggingface.co/collections/robbyant/lingbot-map-v2}}
\newcommand{\methodname}{LingBot-Map~2.0}
\newcommand{\methodnameold}{LingBot-Map}
\newcommand{\methodold}{\texttt{\methodnameold}\xspace}
\newcommand{\method}{\texttt{\methodname}\xspace}

\title{Reconsolidatable Streaming 3D Reconstruction}

\author{%
\begin{center}
    Yihang Chen$^{*}$ \quad
    Lin-Zhuo Chen$^{*}$ \quad
    Jian Gao$^{*}$ \quad
    Jiacheng Dong$^{*}$ \quad
    Yao Yao
    \\[2pt]
    Nan Xue \quad
    Xing Zhu \quad
    Yujun Shen \quad
    Yinghao Xu$^{\dagger}$
    \\[12pt]
    $^{*}$Equal Contribution \qquad
    $^{\dagger}$Project Lead
\end{center}%
}

\begin{document}

\abstract{

In this work, we enhance \methodold with a \textbf{\textit{reconsolidation mechanism}}, yielding \method.
Underlying this design is the observation that popular streaming reconstruction systems process each frame in a single pass, which often causes errors to accumulate over time.
Instead, we adopt a hierarchical strategy in which the lower layers inherit the streaming capability of \methodold, while the upper layers can switch between a streaming mode and a reconsolidation mode.
The two modes share identical weights and differ only in the use of causal versus bidirectional attention.
As a result, reconsolidation can be triggered at any time step, upon which the upper layers jointly re-process all selected keyframes and update the KV cache with the refined results, ensuring that subsequent streaming steps always benefit from the most up-to-date context.
This capability, termed Geometric Context Reconsolidation (GCR), makes drift correction a native part of the model design, obviating the need for offline refinement.
Alongside GCR, we also strengthen the base model of \method, which outperforms existing streaming reconstruction counterparts across nine settings on seven datasets.
Notably, our proposed feed-forward reconsolidation alone, without iterative optimization or an external back-end, suffices to lift \method above loop-closure-equipped streaming methods and offline models on most benchmarks.

}

\maketitle
\justifying

\begin{figure}[H]
\centering
\includegraphics[width=\linewidth,height=.40\textheight,keepaspectratio]{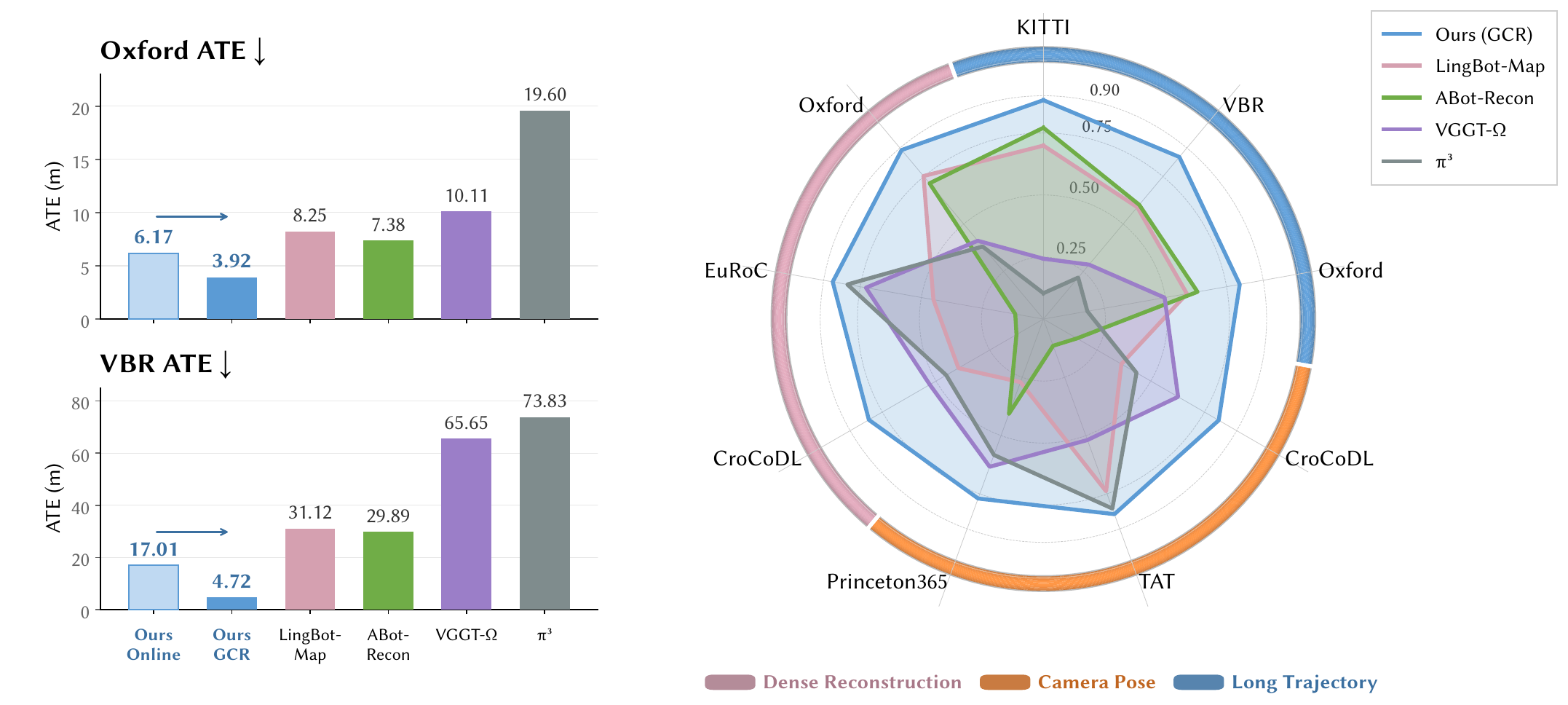}
\captionsetup{skip=3pt}
\caption{\textbf{Comparison of \textnormal{\texttt{\methodname}} with state-of-the-art streaming and offline 3D reconstruction methods.}}
\label{fig:radar}
\end{figure}
\clearpage

\section{Introduction}
\label{sec:intro}

Human spatial memory is formed incrementally, as a continuous stream of visual input is consolidated into a persistent representation of the environment~\cite{mcgaugh2000memory}.
This representation, however, is not fixed once formed.
In cognitive neuroscience, memory reconsolidation refers to the finding that a consolidated memory, upon retrieval, can be updated with new information~\cite{nader2000fear,lee2017update}.
Later observations can therefore revise earlier impressions within the memory itself, so subsequent behavior relies on the updated representation rather than the outdated one.
Can streaming 3D reconstruction models operate similarly, revising their consolidated context as later views arrive?

Streaming 3D reconstruction models~\cite{wang20253dreconstruction,wang2025continuous3d,zhuo2026streaming4d,lan2026stream3rscalable,chen2026ttt3r3d,yuan2026infinitevggtvisual,cheng2026longstreamlong,lu2026ovggto,zhang2026logerlong,cheng2026horizonstreamlong,han2026revisitinglocal,chen2026lingbotmap} consolidate past views into a persistent context, such as a recurrent state or a key-value cache, predicting camera poses and dense geometry for each incoming view.
Since each view enters this context only once and is never revisited, errors from early partial observations accumulate as drift even when later views reveal them.
Correcting them requires revising the consolidated context itself, exposing a tension: later views must reopen it, yet ongoing streaming relies on it as a stable reference.

Existing methods correct streaming reconstructions in different ways, each involving distinct trade-offs.
Hybrid systems such as MASt3R-SLAM~\cite{murai2025mast3rslam} and VGGT-SLAM~\cite{maggio2025vggtslam} couple learned offline 3D models with classical SLAM backends, but their corrections rely on hand-crafted loop detection and iterative pose-graph optimization outside the network.
Streaming models can optionally employ external backends, including the loop-closure stages in HorizonStream~\cite{cheng2026horizonstreamlong} and ABot-Recon~\cite{han2026revisitinglocal} as well as dedicated backends such as CLoSeR~\cite{li2026closer} and Map2SLAM~\cite{gao2026map2slam}, to refine the output trajectory, yet still leave the cached context unchanged, so subsequent views continue to read the drifted context.
SLAM-Former~\cite{yuan2026slamformer} and SLAMFormer-$\infty$~\cite{fang2026slamformerinfinite} revise the model state within a single network, but reopen every layer with full attention, treating per-view local geometry, which later views rarely change, the same as the cross-view pose relations that drift corrupts.
These observations point to a key design principle: revision should act on the consolidated context rather than on its outputs, and it should reopen only the part that later views need to change, keeping the rest fixed as the stable reference that streaming requires, since consecutive views vary gradually unlike unordered image sets (\cref{tab:bidirectional_depth}).

To this end, we introduce \method, a streaming 3D reconstruction model built upon Geometric Context Reconsolidation (GCR).
Analogous to the consolidation of incoming views into a cached geometric context during online inference, GCR reconsolidates this context: once later views are available, it reopens the stored context, updates it jointly across the observed views, and writes the revised context back to the cache.
Following this principle, GCR partitions the context by layer into a \emph{committed context} and a \emph{revisable context}.
The committed context comes from the lower layers, which mainly capture per-view geometry and remain causal, so it is invariant to subsequent views and reused without modification.
Only the revisable context, produced by the upper layers that align views with one another, is recomputed with bidirectional attention among the observed views.
The recomputed context yields revised estimates for past views and updates the cache in place, so subsequent views are predicted from the corrected context.

Because the revisable layers switch between causal and bidirectional attention, we train \method on long sequences with a mixed-mode recipe spanning streaming, reconsolidation, and their transitions.
Reconsolidation can be triggered at any point over ten-thousand-frame streams, suppressing drift and closing loops in one feed-forward pass without place retrieval or pose-graph optimization (\cref{fig:teaser}).

In summary, our contributions are as follows:
\begin{itemize}
\item We introduce \method, built around Geometric Context Reconsolidation (GCR), which revises past estimates by selectively reconsolidating the geometric context that produced them: the committed context is reused as is, and only the revisable context is recomputed.
\item We propose a mixed-mode training recipe that trains the revisable layers for causal streaming, bidirectional reconsolidation, and their transitions, with sequence scale normalization and adjacent relative pose supervision.
\item We evaluate \method on nine settings across seven datasets.
In the streaming setting, it outperforms existing streaming methods on nearly all metrics; after reconsolidation, a purely feed-forward pass without an optimization backend, it surpasses streaming methods with loop-closure backends on long-trajectory and scene-level benchmarks, as well as offline models on most metrics (\cref{fig:radar}).
\end{itemize}

\clearpage
\begingroup
\makeatletter
\setlength{\@fptop}{0pt}
\makeatother
\begin{figure}[p]
\centering
\includegraphics[width=\linewidth]{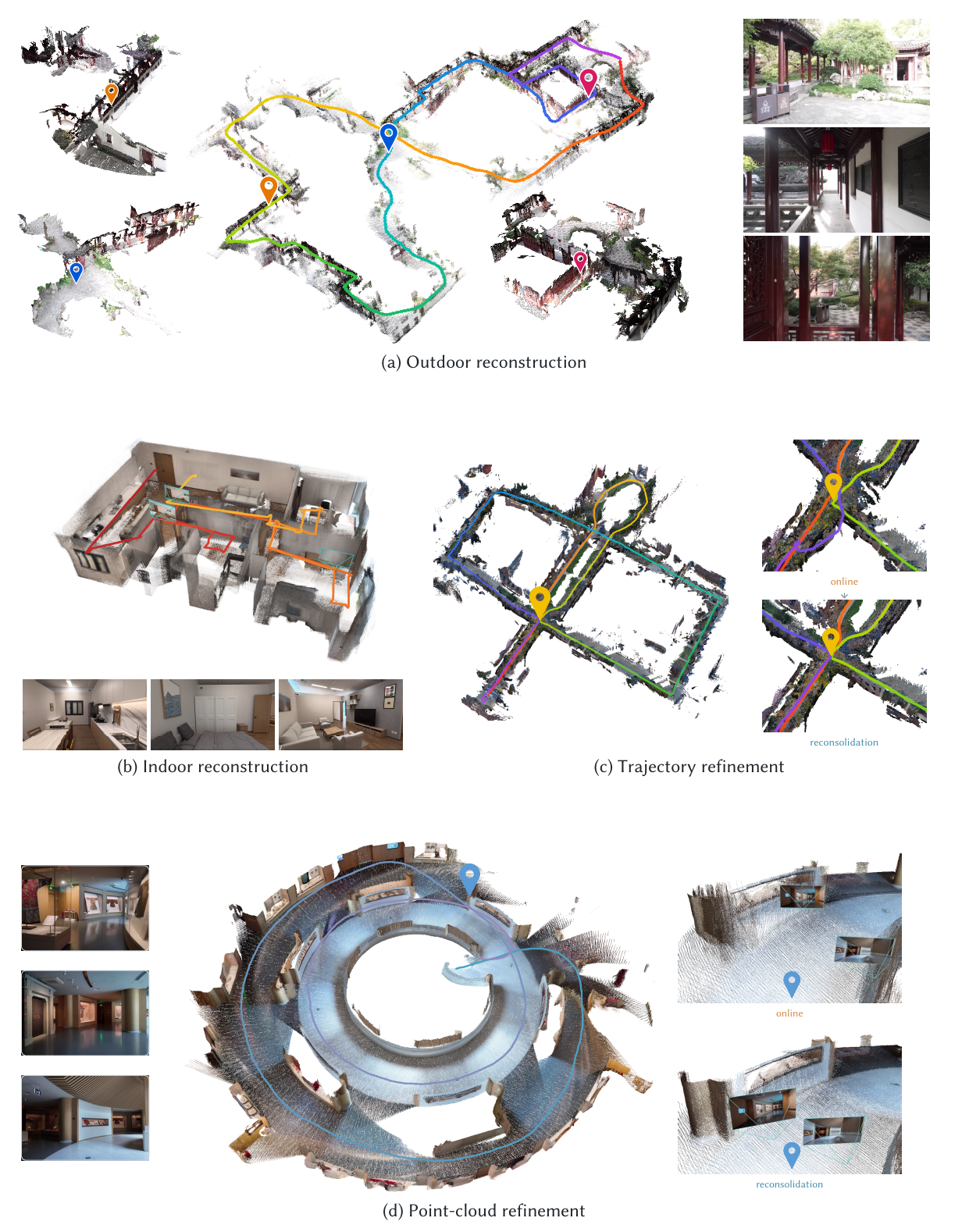}
\captionsetup{skip=4pt}
\caption{\textbf{Reconstruction and reconsolidation with \methodname.} (a) Outdoor reconstruction with three local detail views and (b) indoor reconstruction with its camera trajectory and selected input-camera frusta, both after reconsolidation and shown alongside sampled input frames. (c) Trajectory refinement on a campus revisit and (d) point-cloud refinement in a museum compare online estimates with reconsolidated results.}
\label{fig:teaser}
\end{figure}
\clearpage
\endgroup

\section{Related Work}
\label{sec:related}

\noindent\textbf{Traditional 3D Reconstruction.}
Classical pipelines recover camera motion and dense geometry from correspondences via structure from motion (SfM), visual SLAM, and multiview stereo (MVS)~\cite{schonberger2016structurefrom,campos2021orbslam3,schonberger2016pixelwiseview}, where SfM reconstructs unordered photo collections offline, SLAM tracks video streams online, and MVS estimates dense surfaces from posed views. Learning-augmented systems such as MVSNet~\cite{yao2018mvsnetdepth}, DROID-SLAM~\cite{teed2021droidslam}, and DPV-SLAM~\cite{teed2023deeppatch,lipson2024deeppatch} replace hand-crafted stages with learned cost volumes or dense optical-flow updates, yet accurate reconstruction still hinges on per-scene iterative bundle adjustment and known camera intrinsics.

\noindent\textbf{3D Foundation Models.}
Feed-forward 3D foundation models dispense with per-scene optimization by regressing scene geometry directly from unposed images. DUSt3R~\cite{wang2024dust3rgeometric} predicts pairwise metric pointmaps in a shared coordinate frame but requires post-hoc global alignment beyond two views, whereas VGGT~\cite{wang2025vggtvisual} jointly estimates cameras and dense geometry across arbitrary view sets through alternating frame and multiview attention. Subsequent works advance this paradigm through permutation-equivariant architectures in $\pi^3$~\cite{wang20263permutation}, stronger large-scale geometric accuracy in VGGT-$\Omega$~\cite{wang2026vggt} and Depth Anything 3~\cite{lin2026depthanything}, and token-efficient inference in VGG-T$^3$ and FastVGGT~\cite{elflein2026vggt,shen2026fastvggtfast}. Because these models attend globally over a fixed batch of views, however, memory scales poorly to multi-thousand-frame videos and ongoing streams cannot be updated incrementally without recomputing earlier views.

\noindent\textbf{Streaming 3D Reconstruction.}
For online streams, sequential models instead consolidate incoming views into a persistent state as frames arrive, using fixed-capacity spatial or recurrent memory in Spann3R and CUT3R~\cite{wang20253dreconstruction,wang2025continuous3d} or causal KV caches in StreamVGGT and STream3R~\cite{zhuo2026streaming4d,lan2026stream3rscalable}. To scale to long trajectories with bounded compute, subsequent works explore KV-cache pruning and compression~\cite{yuan2026infinitevggtvisual,lu2026ovggto,cheng2026longstreamlong}, test-time state adaptation in TTT3R and Scal3R~\cite{chen2026ttt3r3d,xie2026scal3rscalable}, sliding-window or chunked relative alignment in LoGeR, HorizonStream, and ABot-Recon~\cite{zhang2026logerlong,cheng2026horizonstreamlong,han2026revisitinglocal}, and compact anchor-window-memory consolidation in LingBot-Map~\cite{chen2026lingbotmap}. Across all of these causal designs, each view enters the context only once and is never revisited, so errors from early partial observations persist in the state and compound into long-range drift.

\noindent\textbf{Global Consistency and Refinement.}
Overcoming such drift calls for revising earlier predictions as new evidence arrives. Classical SLAM enforces global consistency through visual loop detection and pose-graph or bundle adjustment~\cite{campos2021orbslam3,teed2021droidslam}. Building on offline 3D models, hybrid systems such as MASt3R-SLAM and VGGT-SLAM~\cite{murai2025mast3rslam,maggio2025vggtslam,deng2026vggtlong,liu2026suremap} align local submaps via external optimization, while backend stages for streaming models such as CLoSeR and Map2SLAM~\cite{cheng2026horizonstreamlong,han2026revisitinglocal,xie2026scal3rscalable,li2026closer,gao2026map2slam,chen2026poseplaner} refine output trajectories through place retrieval and graph optimization without updating the streaming model's cached context, leaving subsequent frames reading drifted states. Closest to our work, SLAM-Former~\cite{yuan2026slamformer} and SLAMFormer-$\infty$~\cite{fang2026slamformerinfinite} revise states inside a shared transformer, respectively updating cached tokens via global attention and scaling through locally conditioned chunks and iterative graph optimization, yet both reopen every backbone layer and recompute per-view local geometry alongside cross-view alignment. In contrast, \method partitions its context into an invariant \emph{committed context} and an upper \emph{revisable context}, reusing committed layers as is and reopening only revisable layers in one feed-forward pass to correct past estimates and reconsolidate the cache.

\section{Method}
\label{sec:method}

\begin{figure}[!htbp]
\centering
\includegraphics[width=\linewidth]{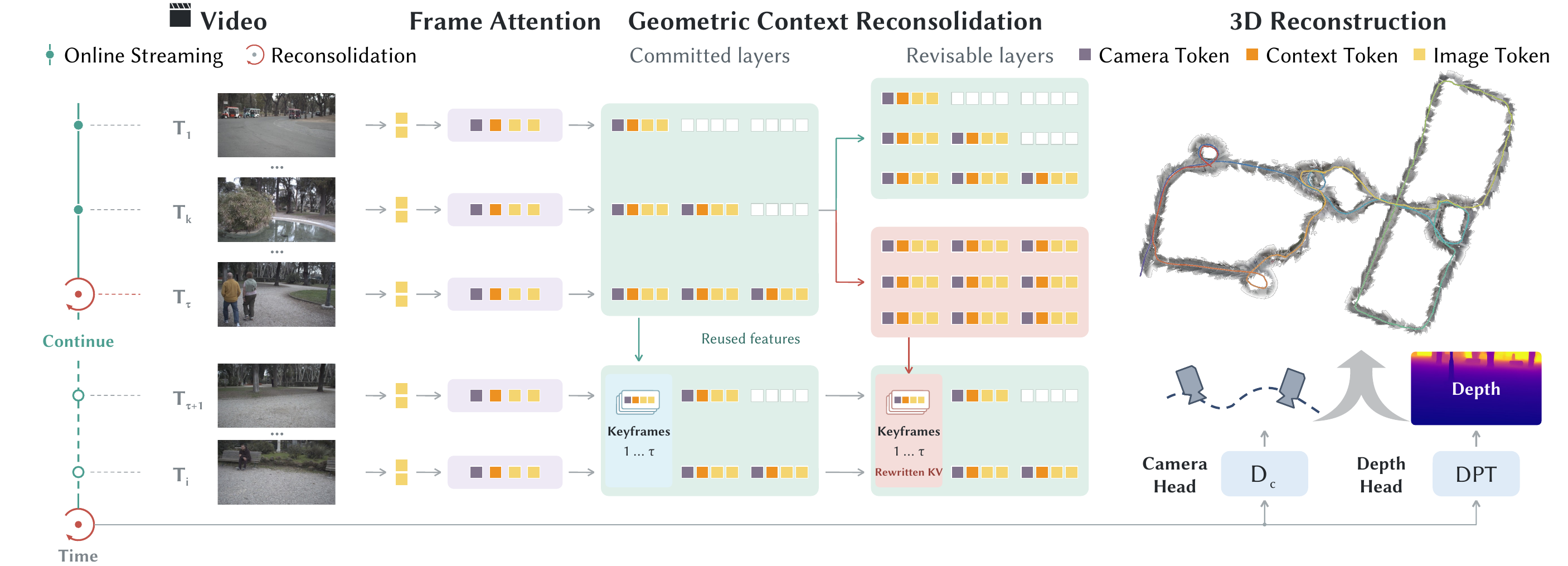}
\caption{\textbf{Method overview.} Reconsolidation reuses the committed context and recomputes the revisable layers bidirectionally from stored tokens. Shared camera and depth heads update past estimates, and subsequent views read the rewritten context. Red arrows mark mid-stream and final reconsolidation; the point cloud shows VBR Pincio after reconsolidation (480 keyframes).}
\label{fig:method}
\end{figure}

\method reconstructs a sequential RGB stream $\{I_t\}$ online, predicting a camera $g_t$ and a depth map $D_t$ for each view while revising earlier predictions as the stream progresses (\cref{fig:method}).
Our key idea is to revise past estimates by reconsolidating the geometric context that produced them, rather than correcting the estimates themselves.
Updating this internal context corrects historical geometry and refreshes the state read by future views at once.
\Cref{sec:prelim} reviews how LingBot-Map consolidates observations into geometric context.
\Cref{sec:split} introduces Geometric Context Reconsolidation, which lets later views correct earlier entries of this context while the causal history that streaming builds on stays intact, \cref{sec:training} presents the full framework and its training, and \cref{sec:inference} describes the inference pipeline.

\subsection{Preliminaries: Geometric Context Attention}
\label{sec:prelim}

We build on the Geometric Context Transformer of LingBot-Map~\cite{chen2026lingbotmap}, which tokenizes each view into image tokens paired with context tokens (a camera token and register tokens), processes them through $L=24$ layers alternating within-view frame attention and cross-view Geometric Context Attention (GCA), and decodes cameras and depth via a causal camera head and a multi-layer DPT~\cite{ranftl2021visiontransformers} head.

\noindent\textbf{Geometric Context Attention.}
GCA keeps cross-view attention causal and compact by drawing on three sources of context.
Streaming retains a subset of frames as keyframes (\cref{sec:inference}).
A fixed initialization prefix $\mathcal A$ establishes a shared coordinate frame and scale, a local window $\mathcal W_t$ of up to $W$ views registers frame $t$ against the most recent keyframes, and the earlier keyframes $\mathcal K_t$ contribute only their context tokens as a long-range trajectory memory.
Following LingBot-Map~\cite{chen2026lingbotmap}, queries from frame $t$ attend to
\begin{equation}
\mathcal V_{\mathrm s}(t)=
{\underbrace{\mathcal T(\mathcal A)}_{\makebox[0pt]{\scriptsize anchors}}}
\cup
{\underbrace{\mathcal T(\mathcal W_t)}_{\makebox[0pt]{\scriptsize window}}}
\cup
{\underbrace{\mathcal C(\mathcal K_t)}_{\makebox[0pt]{\scriptsize memory}}},
\label{eq:causal_visibility}
\end{equation}
where $\mathcal T(\cdot)$ and $\mathcal C(\cdot)$ denote all tokens and context tokens of a frame set.

\noindent\textbf{Consolidation.}
We refer to the streaming key-value (KV) cache as the \emph{geometric context} $\mathcal G_t$, storing the keys and values visible under \cref{eq:causal_visibility} at every GCA layer together with an analogous camera-head cache.
During streaming inference, we maintain this cache through what we term \emph{context consolidation},
\begin{equation}
\mathcal G_t=\mathcal G_{t-1}\oplus\phi_\theta(I_t;\mathcal G_{t-1}),
\label{eq:consolidation}
\end{equation}
where $\phi_\theta$ encodes a new keyframe by attending to $\mathcal G_{t-1}$, and $\oplus$ appends its keys and values while the keyframe leaving the window exposes only its context tokens to later views.
Intermediate frames read $\mathcal G_{t-1}$ without modifying it.
Since consolidation only appends and compresses, each entry is computed once from the views available at the time and never revisited.
Errors arising from these partial observations therefore persist in the context and propagate to every later prediction that attends to it, accumulating as drift over long sequences.

\subsection{Geometric Context Reconsolidation}
\label{sec:split}

\begin{wrapfigure}{R}{0.50\textwidth}
\vspace{-8pt}
\centering
\includegraphics[width=\linewidth]{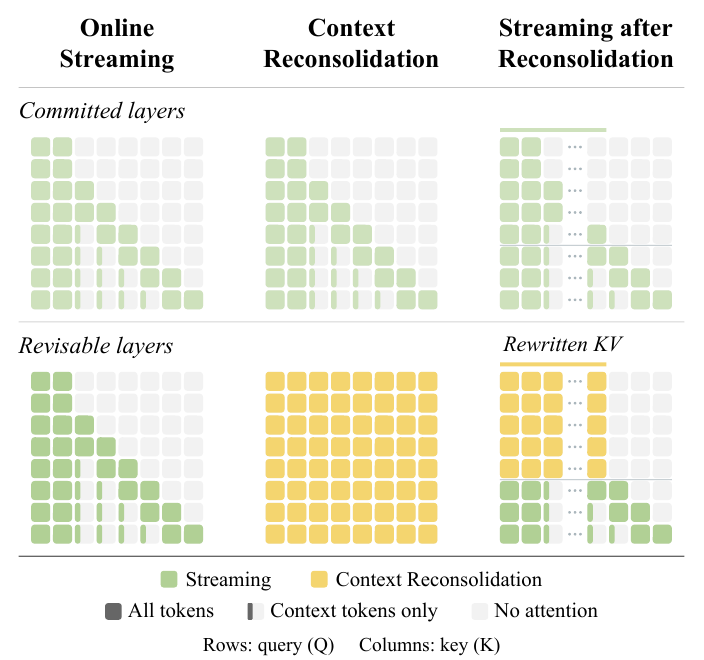}
\captionsetup{skip=4pt,justification=justified,singlelinecheck=false}
\caption{\textbf{Attention across modes.} Committed layers stay causal; revisable layers switch from causal streaming to bidirectional reconsolidation. Streaming then resumes with the rewritten context.}
\label{fig:attention_masks}
\vspace{-6pt}
\end{wrapfigure}

Consolidation thus poses a central challenge for streaming reconstruction: the geometric context must remain fixed to serve as a reference for subsequent views, yet the errors it contains are typically revealed only by those later views.
Following the principle of memory reconsolidation (\cref{sec:intro}), Geometric Context Reconsolidation revises this context itself and reopens only the part that later views mainly need to correct.
It partitions the geometric context by layer into two parts: a \emph{committed context}, which is invariant to subsequent views and serves as a stable reference for continued streaming, and a \emph{revisable context}, which is recomputed with bidirectional attention over the observed keyframes and written back to serve both past estimates and future predictions.
We detail each part below and then describe the three modes that this design enables for streaming reconstruction.

\noindent\textbf{Committed and revisable context.}
The first $L_c$ layers always follow the causal visibility of \cref{eq:causal_visibility}, whether the model is streaming or reconsolidating, and we refer to them as \emph{committed layers}.
The remaining $L-L_c$ \emph{revisable layers} follow the same rule during streaming but attend bidirectionally across the observed keyframes during reconsolidation.
This split reflects that later views mainly correct the relative poses and alignment between views rather than the local geometry of each view: the lower layers mainly capture per-view geometry, whereas the upper layers align views with one another.
We set $L_c=12$ and study this choice in \cref{tab:bidirectional_depth}.
Accordingly, the geometric context at time $\tau$ is split into the committed context $\mathcal G^{\mathrm c}_\tau$, which holds the keys and values of the committed layers, and the revisable context $\mathcal G^{\mathrm r}_\tau$, which holds those of the revisable layers together with the camera head cache, since the camera head reads the output of the revisable layers.

The committed context is invariant to subsequent views.
Intuitively, a committed layer never attends to views that arrive after keyframe $k$, so nothing observed later can alter what it computes for $k$.
Formally, let $F^l_k$ denote the features of keyframe $k$ at a committed layer $l\le L_c$; induction over $l$ shows that $F^l_k$ depends only on frame $k$, the initialization prefix $\mathcal A$, and the keyframes retained before $k$.
Hence neither $F^l_k$ nor $\mathcal G^{\mathrm c}_\tau$ changes when later views arrive or when the revisable layers switch visibility.
In particular, the tokens output by the last committed layer, $H_k=F^{L_c}_k$, are fixed once keyframe $k$ has been processed.

Let $\mathcal K_\tau$ be the keyframes retained up to time $\tau$.
Reconsolidation can therefore be selective, recomputing only $\mathcal G^{\mathrm r}_\tau$ from $H_{\mathcal K_\tau}=\{H_k:k\in\mathcal K_\tau\}$ while reusing $\mathcal G^{\mathrm c}_\tau$ as is:
\begin{equation}
\mathcal G^{\mathrm r}_\tau\leftarrow\rho_\theta(H_{\mathcal K_\tau}),\qquad \mathcal G^{\mathrm c}_\tau\ \text{unchanged},
\label{eq:context_ops}
\end{equation}
where $\rho_\theta$ is described below.

The two parts of the context are maintained and used in three modes that share the same weights (\cref{fig:attention_masks}):
Online Streaming builds both parts, Context Reconsolidation rewrites the revisable part, and Streaming after Reconsolidation reads the two parts together.

\noindent\textbf{Online Streaming.}
Here, each incoming keyframe passes through all $L$ layers under the causal visibility of \cref{eq:causal_visibility}, and, following \cref{eq:consolidation}, the keys and values of its committed layers are appended to $\mathcal G^{\mathrm c}$, while those of its revisable layers and the camera head are appended to $\mathcal G^{\mathrm r}$.
In addition, we store the tokens $H_k$, which are exactly the inputs that the revisable layers need to process this keyframe again, together with the committed-layer features used by the depth head.
These tokens cannot be recovered from the cached keys and values, so they are stored separately; since $H_k$ is fixed, storing them once suffices.

\noindent\textbf{Context Reconsolidation.}
Context Reconsolidation realizes \cref{eq:context_ops} at any time $\tau$: it keeps $\mathcal G^{\mathrm c}_\tau$ and recomputes $\mathcal G^{\mathrm r}_\tau$ from the stored tokens $H_{\mathcal K_\tau}$, so no image is re-encoded and no committed layer is rerun.
Specifically, the revisable layers reprocess the stored tokens of all observed keyframes, now attending bidirectionally across them.
Writing $U_\theta(H;\mathcal V)$ for the revisable layers, which map layer-$L_c$ tokens $H$ to final-layer features under visibility $\mathcal V$, this pass computes
\begin{equation}
Z^{\mathrm r}_{\mathcal K_\tau}=U_\theta\big(H_{\mathcal K_\tau};\mathcal T(\mathcal K_\tau)\big),
\label{eq:refinement_modes}
\end{equation}
where $\mathcal T(\mathcal K_\tau)$ contains all tokens of the observed keyframes.
It uses the same weights as Online Streaming and accesses no frame after $\tau$.
Because every keyframe now attends to all others, earlier keyframes can draw on later views, including revisits.

The new features are used in two ways.
For past estimates, the camera head is replayed over the keyframes in temporal order under its causal rule, decoding $Z^{\mathrm r}_{\mathcal K_\tau}$ into reconsolidated cameras, and the depth head is reapplied to each keyframe using the stored committed-layer features and the new revisable-layer features.
For future views, the keys and values produced in this pass by the revisable layers and the camera head replace those in $\mathcal G^{\mathrm r}_\tau$, yielding $\hat{\mathcal G}^{\mathrm r}_\tau$.

\noindent\textbf{Streaming after Reconsolidation.}
Streaming after Reconsolidation reads the two parts together, resuming from $\mathcal G^{\mathrm c}_\tau\cup\hat{\mathcal G}^{\mathrm r}_\tau$.
For each new view, the committed layers attend to the unchanged causal history in $\mathcal G^{\mathrm c}_\tau$, while the revisable layers and the camera head attend to the rewritten context in $\hat{\mathcal G}^{\mathrm r}_\tau$, and new keyframes are appended to both parts as in Online Streaming.
The two parts stay aligned because both $\hat{\mathcal G}^{\mathrm r}_\tau$ and the new view's inputs to the revisable layers come from the unchanged committed layers, and mixed-mode training (\cref{sec:training}) exposes the revisable layers to causal queries over bidirectionally computed context.
Correcting past predictions alone would leave subsequent views reading stale context, whereas rewriting the context lets their predictions build on the corrected history.

\subsection{\methodname~Framework}
\label{sec:training}

\noindent\textbf{Architecture.}
The overall architecture is illustrated in \cref{fig:method}.
Each input image is encoded by a ViT into image tokens, which are augmented with a camera token and register tokens as in \cref{sec:prelim}.
The tokens are processed by $L=24$ alternating layers of frame attention and GCA, of which the first $L_c=12$ are committed and the remaining $12$ are revisable (\cref{sec:split}).
Frame attention operates within each view, whereas GCA operates across views under \cref{eq:causal_visibility}, or bidirectionally over the observed keyframes in the revisable layers during reconsolidation.
Finally, a camera head refines the final-layer camera token over four iterations with causal attention across views to predict the absolute camera pose, whereas a frame-wise DPT depth head fuses image tokens across both committed and revisable layers (layers 5, 12, 18, and 24) without a cross-view cache to predict the depth map.
All three modes share these weights.

\noindent\textbf{Scale normalization.}
To supervise geometry consistently across modes, before computing the losses we normalize ground truth and predictions separately for each sequence.
Let $\mathcal P$ be a randomly sampled temporal prefix and $M_{\mathcal P}$ its valid ground-truth depth mask.
Using the same prefix and mask for both quantities, we compute
\begin{equation}
s_{\mathrm{gt}}=\operatorname{mean}_{M_{\mathcal P}}D_{\mathrm{gt}},
\qquad
s_{\mathrm{pred}}=\operatorname{mean}_{M_{\mathcal P}}\hat D.
\label{eq:scale_normalization}
\end{equation}
We divide ground-truth depths and camera translations by $s_{\mathrm{gt}}$, and predicted depths and the translations from every camera prediction stage by $s_{\mathrm{pred}}$.
Each scale is shared across the sequence.
Gradients propagate through $s_{\mathrm{pred}}$; rotations and intrinsics are unaffected.
This normalization removes the scale ambiguity at the training loss boundary.
Inference requires no ground truth and reconstructs geometry up to an unknown scale.

\noindent\textbf{Loss function.}
We supervise depth, absolute camera pose, and relative pose after the sequence scale normalization above,
\begin{equation}
\mathcal L=\lambda_{\mathrm{depth}}\mathcal L_{\mathrm{depth}}
+\lambda_{\mathrm{abs}}\mathcal L_{\mathrm{abs}}
+\lambda_{\mathrm{rel}}\mathcal L_{\mathrm{rel}},
\label{eq:training_objective}
\end{equation}
where $\mathcal L_{\mathrm{depth}}$ and $\mathcal L_{\mathrm{abs}}$ follow VGGT and LingBot-Map.
LingBot-Map computes $\mathcal L_{\mathrm{rel}}$ over frame pairs at wide temporal offsets, leaving adjacent pairs as a small fraction of the loss.
Yet errors between consecutive frames directly cause trajectory jitter and local drift.
We instead compute $\mathcal L_{\mathrm{rel}}$ over every adjacent pair.
Given camera poses $T_t\in\mathrm{SE}(3)$ with scale-normalized translations in the world frame, let $\Delta T_t=T_t^{-1}T_{t-1}$ be the ground truth motion increment with rotation $\Delta R_t$ and translation $\Delta\mathbf t_t$, and define $\Delta\hat T_t$ analogously from the predictions.
For a sequence of $N$ frames,
\begin{equation}
\begin{aligned}
\mathcal L_{\mathrm{rel}}=\frac{1}{N-1}\sum_{t=2}^{N}\Big[
&\lambda_{\mathrm{tr}}\big\|\Delta\hat{\mathbf t}_t-\Delta\mathbf t_t\big\|_1\\
&+\lambda_{\mathrm{rot}}\,d_{\mathrm{rot}}(\Delta\hat R_t,\Delta R_t)\Big],
\end{aligned}
\label{eq:relative_loss}
\end{equation}
where $d_{\mathrm{rot}}(\Delta\hat R_t,\Delta R_t)=\arccos\big((\operatorname{tr}(\Delta\hat R_t^\top\Delta R_t)-1)/2\big)$ is the geodesic distance on $\mathrm{SO}(3)$, and $\lambda_{(\cdot)}$ denote scalar loss weights (\cref{sec:streaming_base_training}).
The translation term uses an L1 penalty.
Supervising ground truth increments sharpens local step accuracy without penalizing genuine camera motion, while $\mathcal L_{\mathrm{abs}}$ and $\mathcal L_{\mathrm{depth}}$ supervise cameras and geometry in their normalized coordinate systems.

\noindent\textbf{Mixed-mode training.}
Supporting all three modes with shared weights $\theta$ requires the revisable layers to handle causal context, bidirectional context, and transitions between them.
Continuing from the geometric streaming base model trained under the progressive view curriculum of LingBot-Map~\cite{chen2026lingbotmap} (\cref{sec:base_model_training}), mixed-mode continuous training jointly optimizes consolidation, reconsolidation, and their transitions (\cref{sec:full_training_setup}).
At each batch, we sample one of three attention patterns with equal probability and apply it only to the 12 revisable layers.
The committed layers remain causal under \cref{eq:causal_visibility} in every pattern, including the initialization prefix connections, so the invariance of \cref{sec:split} holds throughout training.
Each pattern exposes these layers to a different kind of context: (i) a causal pattern, where views read consolidated context as in Online Streaming; (ii) a fully bidirectional pattern, as in Context Reconsolidation; and (iii) a bidirectional temporal prefix followed by a causal suffix, where the suffix reads rewritten context as in Streaming after Reconsolidation.
All patterns are implemented by attention masks within a single forward pass, matching the three visibilities in \cref{fig:attention_masks}.
Causal regions use a local pose reference window sampled between 16 and 64 frames, while bidirectional regions attend globally over their permitted views.

\subsection{Inference Pipeline}
\label{sec:inference}

Like autoregressive LLMs, our model caches the keys and values of processed views rather than recomputing them, and we inherit the paged KV-cache runtime of LingBot-Map~\cite{chen2026lingbotmap}.
After jointly encoding the initialization prefix $\mathcal A$, we select one keyframe every $s$ frames (\cref{sec:evaluation_details}): keyframes are consolidated into $\mathcal G$ with $H_k$ stored, whereas intermediate frames are predicted without modifying $\mathcal G$ so every frame receives an online camera and depth.
Context Reconsolidation requires no loop detection and can be triggered at any point, such as after the last frame (\cref{sec:trajectory_results}) or at 75\% of the keyframes (\cref{sec:ablations}) in our experiments.
Since it revises keyframes only, intermediate frames retain their online depths while interpolating cameras from the reconsolidated keyframes, which also form the reconsolidated dense map (\cref{sec:evaluation_details}).

\section{Training \& Inference}
\label{sec:experimental_setup}
\label{sec:training_inference}
Learning both causal online streaming and context reconsolidation within one set of parameters requires staging how sequential memory and cross-mode visibility are acquired.
Directly optimizing cached streaming attention over thousand-frame trajectories from scratch suffers from compounding early-frame drift, while introducing context reconsolidation prematurely prevents the causal layers from establishing a stable online coordinate reference.
We therefore train \method in two main stages: (i) \emph{Geometric Streaming Base Model Training} first builds multi-view 3D perception offline and progressively scales causal Geometric Context Attention over extended trajectories, producing the \emph{geometric streaming base model} of \method (\cref{sec:base_model_training}); and (ii) \emph{Mixed-Mode Continuous Training} continues training this streaming base model across causal, bidirectional, and continuation masks so that a single checkpoint supports Online Streaming, Context Reconsolidation, and Streaming after Reconsolidation (\cref{sec:full_training_setup}).
Data curation and runtime implementation follow in \cref{sec:dataset_details,sec:runtime_details}.

\subsection{Geometric Streaming Base Model Training}
\label{sec:base_model_training}
\label{sec:streaming_base_training}

\noindent\textbf{Multi-view offline training.}
The image encoder is initialized from LingBot-Vision-L~\cite{fu2026vision}, whose local geometric representations outperform DINOv3-L~\cite{simeoni2025dinov3} under strict pose thresholds (\cref{sec:pretraining_initialization}).
Input frames are resized so that the longer side equals 512 pixels under preserved aspect ratios and tokenized on a $16\times 16$ patch grid before entering the 24-layer alternating frame- and cross-view backbone~\cite{wang2025vggtvisual}.
We first train the backbone offline with dense cross-view attention across all views in each batch without causal masking, enabling the network to learn view-order-agnostic correspondence, camera geometry, and depth structure jointly from unstructured photo collections and short video clips.

\noindent\textbf{Long-horizon causal streaming.}
Initialized from this offline multi-view checkpoint, all 24 cross-view layers switch to the causal visibility of \cref{eq:causal_visibility}, registering each incoming view against the initialization anchors, a local patch window, and compact context tokens from earlier keyframes.
Following LingBot-Map~\cite{chen2026lingbotmap}, sequence length expands progressively from short clips to long trajectories so that short-range registration converges before long-range memory dependencies take effect, while the local window size $W$ is drawn from $[16, 64]$ and 16-way context parallelism shards sequence tokens across devices.
Ground-truth and predicted geometry are normalized per sequence via \cref{eq:scale_normalization}, with the scale computed on the sampled prefix and applied to all views and all four camera-head refinement steps (weighted with decay $\gamma=0.6$~\cite{wang2025vggtvisual}), and optimization minimizes \cref{eq:training_objective} with $\lambda_{\mathrm{depth}}=1$, $\lambda_{\mathrm{abs}}=5$, and $\lambda_{\mathrm{rel}}=50$, pairing L1 translation ($\lambda_{\mathrm{tr}}=2$) and $\mathrm{SO}(3)$ geodesic ($\lambda_{\mathrm{rot}}=0.5$) penalties over every consecutive frame pair (\cref{eq:relative_loss}) to suppress local step jitter alongside global trajectory drift.
Completing this curriculum yields the strengthened \emph{geometric streaming base model} of \method for purely causal online reconstruction.

\subsection{Mixed-Mode Continuous Training}
\label{sec:full_training_setup}

\noindent\textbf{Initialization and mixed-visibility sampling.}
Reopening upper causal layers bidirectionally without adaptation severely degrades pose accuracy, as attention projections trained solely on past tokens suffer distribution shift when exposed to future keys and values (\cref{tab:ablation_components}).
Continuing from the geometric streaming base model (\cref{sec:base_model_training}), mixed-mode continuous training starts a fresh optimization schedule using Muon~\cite{liu2025muonscalable} together with AdamW~\cite{loshchilov2019decoupledweight} at a peak learning rate of $5\times 10^{-5}$ under 16-way context parallelism.
Across every batch, the lower $L_c=12$ committed layers remain strictly causal under \cref{eq:causal_visibility} so their cached per-view tokens $H_k$ stay invariant, while the 12 upper revisable layers sample one of three attention masks with equal probability (\cref{sec:training,fig:attention_masks}): a causal mask (preserving Online Streaming with $W\in[16,64]$), a global bidirectional mask over all views in the sequence (learning Context Reconsolidation on top of fixed $H_k$), or a bidirectional prefix followed by a causal suffix (simulating Streaming after Reconsolidation, where subsequent queries attend to the rewritten cache $\hat{\mathcal G}^{\mathrm r}_\tau$).

\noindent\textbf{Unified cross-mode supervision.}
Because the camera and DPT depth heads are shared across all three modes, representations produced under causal, bidirectional, and prefix-to-causal visibilities must remain geometrically compatible.
We therefore apply the same sequence-level scale normalization (\cref{eq:scale_normalization}) and loss weights as in \cref{sec:base_model_training} across all three mask types, producing the single shared checkpoint evaluated throughout \cref{sec:experiments}.

\subsection{Training Data and Sampling}
\label{sec:dataset_details}

\noindent\textbf{Dataset overview.}
\Cref{tab:dataset_composition} summarizes the 36 real and synthetic sources used for training, covering indoor rooms, outdoor streets, aerial scenes, and object-centric captures while excluding every evaluation benchmark in \cref{sec:benchmark}.
Static multi-view collections and short clips serve offline training, whereas extended video trajectories support both causal streaming and mixed-mode continuous training.

\noindent\textbf{Offline training mixture.}
Offline training uses unstructured multi-view datasets, including BlendedMVS~\cite{yao2020blendedmvslarge}, MegaDepth~\cite{li2018megadepthlearning}, MegaDepthX~\cite{li2026longtail}, and AerialMegaDepth~\cite{vuong2025aerialmegadepthlearning}, together with 3D object collections Co3D~\cite{reizenstein2021commonobjects}, Objaverse~\cite{deitke2023objaverseuniverse}, and Texverse~\cite{zhang2025texverseuniverse}, as well as indoor and synthetic video sources including ScanNet~\cite{dai2017scannetrichly}, ScanNet++~\cite{yeshwanth2023scannethigh}, WildRGBD~\cite{xia2024rgbdobjects}, Aria Synthetic~\cite{avetisyan2024scenescriptreconstructing}, ADT~\cite{pan2023ariadigital}, PointOdyssey~\cite{zheng2023pointodysseylarge}, and Kubric~\cite{greff2022kubricscalable}.
This heterogeneous mix exposes the backbone to wide viewpoint baselines, diverse focal lengths, and fine object geometry.

\noindent\textbf{Causal and mixed-mode mixture.}
For causal streaming and mixed-mode continuous training, unordered photo collections are dropped so every training sample reflects continuous camera motion (\cref{tab:dataset_composition}).
This trajectory pool concentrates on multi-room indoor navigation rendered with Habitat~\cite{savva2019habitatplatform} across HM3D~\cite{ramakrishnan2021habitatmatterport}, Matterport3D~\cite{chang2017matterport3dlearning}, and Gibson~\cite{xia2018gibsonenv}, alongside large-scale outdoor, driving, aerial, and game-engine streams from TartanAir/TartanAir2~\cite{wang2020tartanairdataset}, TartanGround~\cite{patel2025tartangroundlarge}, MidAir~\cite{fonder2019midair}, MatrixCity~\cite{li2023matrixcitylarge}, KITTI-360~\cite{liao2023kitti360}, and the OmniWorld-Game split of OmniWorld~\cite{zhou2026omniworldmulti}.

\noindent\textbf{Trajectory sampling.}
Ordered video clips for causal and mixed-mode training are drawn from random starting points with variable frame strides; whenever a sampled walk reaches the end of a recording before filling the target length, traversal reflects backward from the boundary~\cite{chen2026lingbotmap} to form an unbroken pass with natural revisit overlap.
The resulting sequences supervise causal tracking in the streaming base stage and, paired with the three visibility masks above, train context reconsolidation and post-reconsolidation continuation during mixed-mode continuous training.

\begin{table}[!htbp]
\centering
\caption{\textbf{Training dataset composition.} Modality indicates unstructured multi-view sets vs.\ ordered video sequences. Offline train denotes multi-view offline training; Causal/mix denotes long-sequence causal streaming and mixed-mode continuous training. On OmniWorld, causal/mix training uses the OmniWorld-Game subset; HM3D, Matterport3D, and Gibson are rendered with Habitat~\cite{savva2019habitatplatform}.}
\label{tab:dataset_composition}
\footnotesize
\setlength{\tabcolsep}{1.5pt}
\renewcommand{\arraystretch}{1.30}
\begin{minipage}[t]{0.35\textwidth}\vspace{0pt}
\centering
\begin{tabular*}{\linewidth}{@{\extracolsep{\fill}}llcc@{}}
\toprule
Dataset & Modality & \shortstack{Offline\\train} & \shortstack{Causal/\\mix} \\
\midrule
BlendedMVS~\cite{yao2020blendedmvslarge} & Multi-view & \ding{51} & \ding{51} \\
HyperSim~\cite{roberts2021hypersimphotorealistic} & Multi-view & \ding{51} & \ding{51} \\
MegaDepth~\cite{li2018megadepthlearning} & Multi-view & \ding{51} & -- \\
MegaDepthX~\cite{li2026longtail} & Multi-view & \ding{51} & -- \\
AerialMegaDepth~\cite{vuong2025aerialmegadepthlearning} & Multi-view & \ding{51} & -- \\
MVS-Synth~\cite{huang2018deepmvslearning} & Multi-view & \ding{51} & \ding{51} \\
GTA-SfM~\cite{wang2020flowmotion} & Multi-view & \ding{51} & \ding{51} \\
Co3D~\cite{reizenstein2021commonobjects} & Multi-view & \ding{51} & -- \\
Objaverse~\cite{deitke2023objaverseuniverse} & Multi-view & \ding{51} & -- \\
Texverse~\cite{zhang2025texverseuniverse} & Multi-view & \ding{51} & -- \\
ScanNet~\cite{dai2017scannetrichly} & Video & \ding{51} & \ding{51} \\
ScanNet++~\cite{yeshwanth2023scannethigh} & Video & \ding{51} & \ding{51} \\
\bottomrule
\end{tabular*}
\end{minipage}\hfill
\begin{minipage}[t]{0.335\textwidth}\vspace{0pt}
\centering
\begin{tabular*}{\linewidth}{@{\extracolsep{\fill}}llcc@{}}
\toprule
Dataset & Modality & \shortstack{Offline\\train} & \shortstack{Causal/\\mix} \\
\midrule
SceneNet RGB-D~\cite{mccormac2017scenenetrgb} & Video & \ding{51} & \ding{51} \\
WildRGBD~\cite{xia2024rgbdobjects} & Video & \ding{51} & \ding{51} \\
UnrealStereo4K~\cite{tosi2021smdnets} & Video & \ding{51} & \ding{51} \\
Aria Synthetic~\cite{avetisyan2024scenescriptreconstructing} & Video & \ding{51} & \ding{51} \\
ADT~\cite{pan2023ariadigital} & Video & \ding{51} & \ding{51} \\
DL3DV~\cite{ling2024dl3dv10k} & Video & \ding{51} & \ding{51} \\
DynamicReplica~\cite{karaev2023dynamicstereoconsistent} & Video & \ding{51} & \ding{51} \\
TartanAir~\cite{wang2020tartanairdataset} & Video & \ding{51} & \ding{51} \\
TartanAir2~\cite{wang2020tartanairdataset} & Video & \ding{51} & \ding{51} \\
TartanGround~\cite{patel2025tartangroundlarge} & Video & \ding{51} & \ding{51} \\
Virtual KITTI 2~\cite{cabon2020virtualkitti} & Video & \ding{51} & \ding{51} \\
KITTI-360~\cite{liao2023kitti360} & Video & -- & \ding{51} \\
\bottomrule
\end{tabular*}
\end{minipage}\hfill
\begin{minipage}[t]{0.308\textwidth}\vspace{0pt}
\centering
\begin{tabular*}{\linewidth}{@{\extracolsep{\fill}}llcc@{}}
\toprule
Dataset & Modality & \shortstack{Offline\\train} & \shortstack{Causal/\\mix} \\
\midrule
MatrixCity~\cite{li2023matrixcitylarge} & Video & \ding{51} & \ding{51} \\
MidAir~\cite{fonder2019midair} & Video & \ding{51} & \ding{51} \\
MapFree~\cite{arnold2022mapfree} & Video & \ding{51} & \ding{51} \\
SHIFT~\cite{sun2022shiftsynthetic} & Video & \ding{51} & \ding{51} \\
OmniWorld~\cite{zhou2026omniworldmulti} & Video & \ding{51} & \ding{51} \\
PointOdyssey~\cite{zheng2023pointodysseylarge} & Video & \ding{51} & -- \\
Kubric~\cite{greff2022kubricscalable} & Video & \ding{51} & -- \\
Robo3R~\cite{yang2026robo3renhancing} & Video & \ding{51} & -- \\
HM3D~\cite{ramakrishnan2021habitatmatterport} & Video & -- & \ding{51} \\
Matterport3D~\cite{chang2017matterport3dlearning} & Video & -- & \ding{51} \\
Gibson~\cite{xia2018gibsonenv} & Video & -- & \ding{51} \\
Internal Data & Video & \ding{51} & \ding{51} \\
\bottomrule
\end{tabular*}
\end{minipage}
\end{table}

\subsection{Inference and Cache Implementation}
\label{sec:runtime_details}

\noindent\textbf{Paged KV-cache streaming.}
Building on the paged KV-cache runtime of LingBot-Map~\cite{chen2026lingbotmap}, inference first encodes the initialization prefix $\mathcal A$ jointly to establish a shared coordinate frame and scale, and then processes incoming frames sequentially at the keyframe stride defined in \cref{sec:benchmark}.
Image-patch keys and values occupy fixed-size pages for $\mathcal A$, the local pose-reference window $\mathcal W_t$, and a single scratch slot, whereas older keyframes $\mathcal K_t$ retain only their compact context tokens in packed pages.
Consolidating a keyframe appends its keys and values to the persistent context, recycles the patch-token page of the keyframe leaving $\mathcal W_t$ in $O(1)$ time, and archives its layer-$L_c$ tokens $H_k$ together with committed DPT features from layers 5 and 12 once to host memory.
Intermediate non-keyframes are decoded online against the active cache via the scratch page, which is immediately rolled back so that persistent state grows only at keyframes.

\begin{wrapfigure}{R}{0.50\textwidth}
\vspace{-8pt}
\centering
\captionsetup{skip=4pt,justification=raggedright,singlelinecheck=false}
\includegraphics[width=\linewidth]{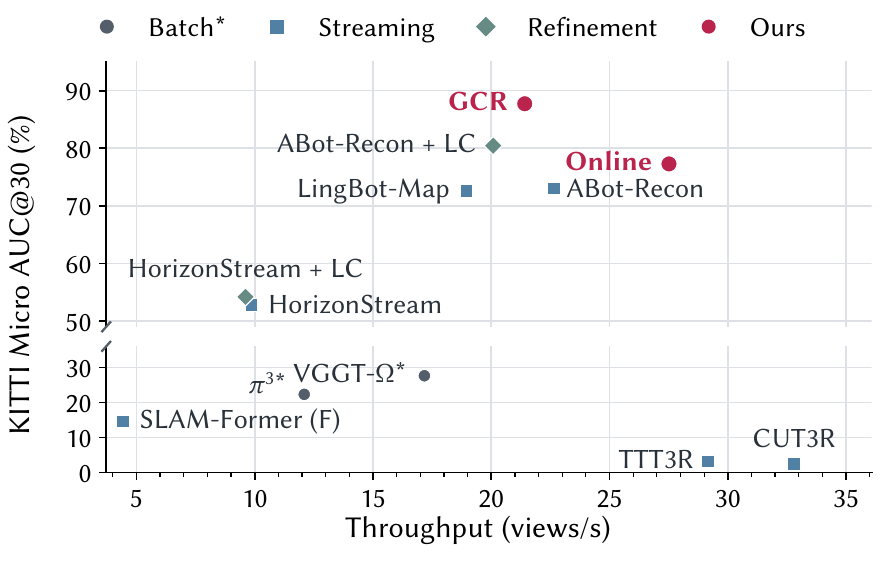}
\caption{\textbf{Accuracy--speed trade-off on KITTI.} Eleven-sequence AUC@30 versus single-GPU throughput. $^{*}$: offline keyframe inference; GCR: reconsolidation of cached keyframes. The vertical axis omits $36\text{--}49\%$.}
\label{fig:kitti_speed_tradeoff}
\vspace{-6pt}
\end{wrapfigure}

\noindent\textbf{Context reconsolidation and stream continuation.}
Context Reconsolidation can be invoked at any point in the stream without a loop detector---such as at sequence end or mid-stream at 75\% of the keyframes in our evaluations---by retrieving the archived $H_k$ and committed DPT features from host memory and recomputing only the 12 revisable layers bidirectionally across the retained keyframes.
The shared camera and depth heads decode updated keyframe trajectories and depth maps (\cref{sec:inference,sec:evaluation_details}), while the recomputed revisable-layer and camera-head KV pools replace $\mathcal G^{\mathrm r}_\tau$ in place and reuse the invariant committed cache $\mathcal G^{\mathrm c}_\tau$ unchanged.
Subsequent incoming frames then attend directly to $\mathcal G^{\mathrm c}_\tau \cup \hat{\mathcal G}^{\mathrm r}_\tau$, so reconsolidation simultaneously corrects historical estimates and refreshes the state used for future predictions.

\noindent\textbf{Accuracy--speed comparison.}
\label{sec:accuracy_speed}
We benchmark inference speed on a single GPU using the 1,101-frame KITTI~\texttt{07} sequence after warmup, excluding model loading and disk I/O while including loop-closure stages where applicable, and plot throughput against 11-sequence KITTI AUC@30 in \cref{fig:kitti_speed_tradeoff}.

Online streaming and loop-closure baselines process all 1,101 raw frames. Fixed-state recurrent models such as CUT3R and TTT3R achieve high raw frame rates because their recurrent state has constant size, but suffer severe trajectory drift over kilometer-scale driving. Among causal methods that preserve long-horizon geometry, Ours (Online) runs in real time at 27.51~views/s while achieving 77.28 AUC@30, outperforming LingBot-Map, ABot-Recon, and HorizonStream in both speed and accuracy by recycling evicted patch-token pages in constant time.

Offline batch models and the SLAM-Former frontend are timed on the 373 retained keyframes. Starting from an existing causal cache over the same 373 retained keyframes without counting cache construction, Context Reconsolidation runs at 21.41~cached keyframes/s---reusing the committed layers without re-encoding images---and raises AUC@30 to 87.73, surpassing external loop-closure pipelines without place retrieval or pose-graph optimization. Combining full-stream online inference with sequence-end reconsolidation corresponds to an end-to-end throughput of 19.17~frames/s across the full sequence.

\section{Evaluation Benchmark}
\label{sec:benchmark}

Streaming 3D reconstruction must generalize from room-scale interiors and object-scale scans to multi-kilometer urban traversals under varying motion patterns, revisit intervals, and illumination conditions.
To evaluate both online consolidation and context reconsolidation across these regimes, we establish an evaluation benchmark spanning nine settings across seven primary datasets, described in \cref{sec:eval_datasets} together with the evaluation protocols and metrics in \cref{sec:evaluation_details}.

\subsection{Datasets}
\label{sec:eval_datasets}

\begingroup
\addtolength{\parskip}{4pt}
\noindent\textbf{VBR}~\cite{brizi2024vbrvision} is a large-scale vision benchmark captured across historic urban streets and university campuses in Rome, featuring multi-kilometer loops, wide-baseline revisits, and centimeter-accurate RTK/LiDAR-inertial ground-truth trajectories.
We evaluate on all seven sequences with public ground-truth trajectories: Colosseo-0, Campus-0, Campus-1, Pincio-0, Spagna-0, Diag-0, and Ciampino-1, spanning 8,815 to 18,846 frames and up to 5.20\,km per trajectory, for both long-trajectory estimation and revisit consistency.

\noindent\textbf{Oxford Spires}~\cite{tao2026oxfordspires} is a campus-scale dataset captured across historic Oxford courtyards, narrow passages, and indoor-outdoor transitions, providing millimeter-accurate LiDAR-inertial camera trajectories and laser-scanned reference geometry.
Following LingBot-Map~\cite{chen2026lingbotmap}, we use all 10 sequences with ground-truth trajectories: Bodleian Library-02, Christ Church-02, 03, and 05, Keble College-02 through 05, and Observatory Quarter-01 and 02, ranging from 3,961 to 13,518 frames.
We evaluate full-sequence tracking across all valid frames, dense 3D reconstruction at stride $s=10$, and visual revisit alignment.

\noindent\textbf{KITTI Odometry}~\cite{geiger2012weready} provides city-scale autonomous driving trajectories across urban streets, rural roads, and highways.
We evaluate long-trajectory tracking across all 11 sequences with ground-truth poses, namely \texttt{00} through \texttt{10}, spanning 271 to 4,661 frames and up to 5.10\,km, and evaluate loop-closure consistency on the six sequences containing revisits: \texttt{00}, \texttt{02}, \texttt{05}, \texttt{06}, \texttt{07}, and \texttt{09}.

\noindent\textbf{CroCoDL}~\cite{blum2025crocodlcross} is a cross-device collaborative localization and mapping benchmark registered against NavVis LiDAR-scanned 3D reference models, spanning unstructured indoor and outdoor ARCHE environments---including dimly lit basements, semi-collapsed structures, open-sky ruins, and outdoor plazas---captured under varying illumination and scene layouts.
After excluding two sequences with inconsistent image-size metadata or mixed orientations, we evaluate both camera pose estimation and dense 3D reconstruction on 62 handheld sequences sampled from frame~0 with stride $s=5$.

\noindent\textbf{Tanks and Temples}~\cite{knapitsch2017tanksand} is a multi-view benchmark of indoor and outdoor large-scale scenes and structures with laser-scanned ground truth.
Following LingBot-Map~\cite{chen2026lingbotmap}, we evaluate camera pose estimation on six scenes---Barn, Caterpillar, Church, Ignatius, Meetingroom, and Truck---using all 2,185 images at stride $s=1$.

\noindent\textbf{Princeton365}~\cite{kayan2025princeton365diverse} is a diverse RGB video benchmark spanning object-scanning, indoor, and outdoor scenes, with ground-truth six-degree-of-freedom camera poses calibrated via a synchronized $360^\circ$ camera rig and multi-board markers.
We evaluate camera pose estimation across all 57 validation scanning sequences, sampling from frame~0 with stride $s=10$.

\noindent\textbf{EuRoC Vicon}~\cite{burri2016eurocmicro} comprises the six indoor Vicon Room micro-aerial-vehicle sequences from the EuRoC MAV benchmark, featuring rapid six-degree-of-freedom motion paired with motion-capture poses and millimeter-accurate Leica 3D laser scans.
We evaluate dense 3D reconstruction across all six sequences sampled with stride $s=5$.

\subsection{Evaluation Protocols and Metrics}
\label{sec:evaluation_details}
\label{sec:scene_pose_protocol}
\label{sec:dense_protocol}

\noindent\textbf{Long Trajectory Estimation.}
Across VBR, Oxford Spires, and KITTI, uniform keyframe sampling~\cite{chen2026lingbotmap} retains the first eight frames for $\mathcal A$ and selects subsequent frames with stride $s = \lceil N/B \rceil$, using $B=480$ in the main tables and $B=320$ in lighter ablations.
After Context Reconsolidation, non-keyframe camera centers are linearly interpolated and rotations in \cref{tab:kitti_vbr_summary} are recovered from the nearest retained keyframe $k$ via $R_i^{\mathrm r} = R_k^{\mathrm r}(R_k^{\mathrm s})^{-1}R_i^{\mathrm s}$.
Streaming baselines other than SLAM-Former process the full RGB stream directly, as does Scal3R~\cite{xie2026scal3rscalable} on the shorter KITTI sequences via its official disk offloading.
By contrast, offline models including VGGT-$\Omega$~\cite{wang2026vggt}, $\pi^3$~\cite{wang20263permutation}, and DA3~\cite{lin2026depthanything}, together with SLAM-Former~\cite{yuan2026slamformer} and Scal3R on the longer VBR and Oxford Spires streams, exceed GPU or host memory on multi-thousand-frame inputs; we therefore evaluate them on the same uniformly selected keyframes and recover intermediate frames via linear camera-center interpolation and SLERP~\cite{shoemake1985animatingrotation}.
All methods are scored over the full timeline with valid ground truth using sequence-averaged Sim(3)-aligned Absolute Trajectory Error (ATE, in meters) and AUC@5 and AUC@30 over all valid within-sequence camera pairs.
For KITTI ablations, we additionally report sequence-averaged translational Relative Pose Error (RPE-t, in meters), computed as RMSE over consecutive full-timeline frame pairs after Sim(3) alignment.
For layer-split ablations, depth Absolute Relative Error (AbsRel) is averaged over valid pixels, evaluated keyframes, and sequences after per-sequence median scale alignment.

\noindent\textbf{Camera Pose Estimation.}
On CroCoDL, Tanks and Temples, and Princeton365, poses are scored directly on the sampled views without interpolating skipped frames.
We report AUC@15 and AUC@30 by integrating the cumulative accuracy of the per-pair maximum of relative rotation and translation angular errors up to thresholds of $15^\circ$ and $30^\circ$, weighted by valid within-sequence camera-pair counts, alongside sequence-averaged Sim(3)-aligned ATE.

\noindent\textbf{Dense Reconstruction.}
On CroCoDL, EuRoC Vicon, and Oxford Spires, reconstructions are evaluated directly on the sampled views without depth interpolation or confidence filtering.
Predicted and reference point clouds are aligned via Umeyama Sim(3)~\cite{umeyama1991leastsquares} followed by rigid ICP~\cite{besl1992methodregistration} within the visible region shared across methods, and uniformly voxel-downsampled before computing Accuracy (Acc), Completeness (Comp), and F1 score.
We use an F1 threshold of $0.25$\,m on CroCoDL and EuRoC Vicon, and $4$\,m on Oxford Spires.

\noindent\textbf{Revisit Consistency.}
Following NetVLAD~\cite{arandjelovic2016netvladcnn}, detector-independent ground-truth revisits are defined by matching each view to its spatially nearest earlier camera within 5\,m and $45^\circ$ after at least 500 frames and 100\,m of travel; matches separated by at most 20 frames are merged into a single revisit event.
Under full-trajectory Sim(3) alignment, we report whole-trajectory ATE alongside the local translational error $e_t$ in meters and rotational error $e_R$ in degrees of the relative pose $T_i^{-1}T_j$ across matched revisit pairs.
\par
\endgroup

\begingroup
\section{Experiments}
\label{sec:experiments}

We compare \method with state-of-the-art streaming, global-refinement, and offline methods across pose and trajectory estimation (\cref{sec:trajectory_results}), dense 3D reconstruction (\cref{sec:small_scene_results}), ablation studies (\cref{sec:ablations}), and revisit consistency (\cref{sec:revisit_evaluation}), following the training setup in \cref{sec:experimental_setup} and benchmark protocols in \cref{sec:benchmark}.
Unless otherwise specified, all results---including both Ours (Online) and Ours (Reconsolidation)---are evaluated using the single final checkpoint from mixed-mode continuous training (\cref{sec:full_training_setup}).

\subsection{Pose and Trajectory Estimation}
\label{sec:trajectory_results}

Across long-trajectory benchmarks (\cref{sec:long_trajectory_results}) and scene-level camera pose benchmarks (\cref{sec:camera_pose_results}), we report pose AUC over all valid within-sequence camera pairs alongside sequence-averaged Sim(3)-aligned ATE for Online Streaming and Context Reconsolidation.

\makeatletter
\setlength{\@fpsep}{4pt plus 2fil}
\makeatother
\begin{table}[!p]
\centering
\captionsetup{skip=3pt}
\caption{\textbf{Long-trajectory estimation.} AUC@5 and AUC@30 (\%) over valid within-sequence camera pairs and sequence-averaged Sim(3)-aligned ATE (m). Within each group, bold/underline mark best/second best; across all methods, blue/pale blue mark best/second best (ties share color). LC denotes loop closure. Non-keyframe poses linearly interpolate camera centers, with SLERP~\cite{shoemake1985animatingrotation} ($^{\dagger}$) or nearest-keyframe relative online rotation ($^{\ddagger}$); Scal3R~\cite{xie2026scal3rscalable} runs on all KITTI frames.}
\label{tab:kitti_vbr_summary}
\small
\setlength{\tabcolsep}{2.5pt}
\renewcommand{\arraystretch}{1.0}
\begin{adjustbox}{max width=\textwidth}
\begin{tabular}{@{}clccccccccc@{}}
\toprule
 &  & \multicolumn{3}{c}{VBR} & \multicolumn{3}{c}{Oxford Spires} & \multicolumn{3}{c}{KITTI} \\
\cmidrule(lr){3-5}\cmidrule(lr){6-8}\cmidrule(lr){9-11}
LC & Method & AUC@5$\uparrow$ & AUC@30$\uparrow$ & ATE$\downarrow$ & AUC@5$\uparrow$ & AUC@30$\uparrow$ & ATE$\downarrow$ & AUC@5$\uparrow$ & AUC@30$\uparrow$ & ATE$\downarrow$ \\
\midrule
\multicolumn{11}{@{}l}{\textit{Offline}} \\
\ding{55} & VGGT-$\Omega$$^{\dagger}$ & \textbf{4.91} & \textbf{38.41} & \textbf{65.65} & \textbf{14.59} & \textbf{63.96} & \textbf{10.11} & \textbf{4.23} & \textbf{27.61} & \textbf{122.19} \\
\ding{55} & $\pi^{3}$$^{\dagger}$ & 2.36 & 24.95 & 73.83 & 5.78 & 37.44 & 19.60 & 2.94 & 22.30 & 166.69 \\
\ding{55} & $\pi^{3}$X$^{\dagger}$ & \underline{4.49} & \underline{32.90} & \underline{66.16} & 10.04 & \underline{51.09} & \underline{13.23} & \underline{3.82} & \underline{26.03} & \underline{155.29} \\
\ding{55} & DA3$^{\dagger}$ & 2.20 & 23.48 & 81.69 & \underline{11.17} & 45.99 & 19.33 & 1.60 & 15.31 & 183.50 \\
\midrule
\multicolumn{11}{@{}l}{\textit{Streaming}} \\
\ding{55} & SLAM-Former (frontend)$^{\dagger}$ & 0.68 & 13.66 & 87.28 & 1.16 & 14.28 & 28.81 & 0.78 & 14.57 & 199.42 \\
\ding{55} & LingBot-Map & \underline{8.66} & \underline{50.74} & 31.12 & \underline{29.85} & \underline{68.23} & 8.25 & 17.24 & 72.59 & 24.07 \\
\ding{55} & HorizonStream & 5.51 & 42.52 & \underline{29.21} & 1.09 & 20.37 & 18.39 & 6.52 & 52.81 & 22.50 \\
\ding{55} & ABot-Recon & 5.10 & 42.46 & 29.89 & 5.13 & 46.83 & \underline{7.38} & \underline{21.77} & \underline{73.04} & \underline{18.32} \\
\ding{55} & \textbf{Ours (Online)} & \textbf{14.74} & \textbf{63.57} & \textbf{17.01} & \cellcolor[HTML]{E8F1FA}\textbf{43.41} & \textbf{78.59} & \textbf{6.17} & \textbf{22.42} & \textbf{77.28} & \textbf{17.56} \\
\midrule
\multicolumn{11}{@{}l}{\textit{Global refinement}} \\
\ding{55} & SLAM-Former (w/ backend)$^{\dagger}$ & 1.04 & 15.07 & 87.41 & 4.14 & 26.54 & 26.56 & 0.91 & 13.56 & 211.62 \\
\ding{51} & CLoSeR & \cellcolor[HTML]{E8F1FA}\underline{17.76} & \cellcolor[HTML]{E8F1FA}\underline{71.30} & 11.16 & \underline{29.94} & \cellcolor[HTML]{E8F1FA}\underline{81.16} & \cellcolor[HTML]{E8F1FA}\underline{4.45} & 21.07 & 79.20 & \cellcolor[HTML]{E8F1FA}\underline{12.79} \\
\ding{51} & Scal3R w/ LC$^{\dagger}$ & 2.14 & 15.73 & 60.22 & 11.33 & 49.92 & 11.08 & \cellcolor[HTML]{E8F1FA}\underline{39.85} & \cellcolor[HTML]{E8F1FA}\underline{84.95} & 13.72 \\
\ding{51} & HorizonStream w/ LC & 4.56 & 43.98 & 17.78 & 1.21 & 18.97 & 15.32 & 6.06 & 54.18 & 17.98 \\
\ding{51} & ABot-Recon w/ LC & 10.92 & 64.80 & \cellcolor[HTML]{E8F1FA}\underline{9.98} & 5.82 & 58.72 & 5.37 & 24.77 & 80.45 & 13.56 \\
\ding{55} & \textbf{Ours (Reconsolidation)}$^{\ddagger}$ & \cellcolor[HTML]{BFD9F2}\textbf{37.94} & \cellcolor[HTML]{BFD9F2}\textbf{85.47} & \cellcolor[HTML]{BFD9F2}\textbf{4.72} & \cellcolor[HTML]{BFD9F2}\textbf{74.71} & \cellcolor[HTML]{BFD9F2}\textbf{93.16} & \cellcolor[HTML]{BFD9F2}\textbf{3.92} & \cellcolor[HTML]{BFD9F2}\textbf{41.00} & \cellcolor[HTML]{BFD9F2}\textbf{87.73} & \cellcolor[HTML]{BFD9F2}\textbf{9.42} \\
\bottomrule
\end{tabular}
\end{adjustbox}
\end{table}

\begin{figure}[!p]
\centering
\includegraphics[width=\linewidth]{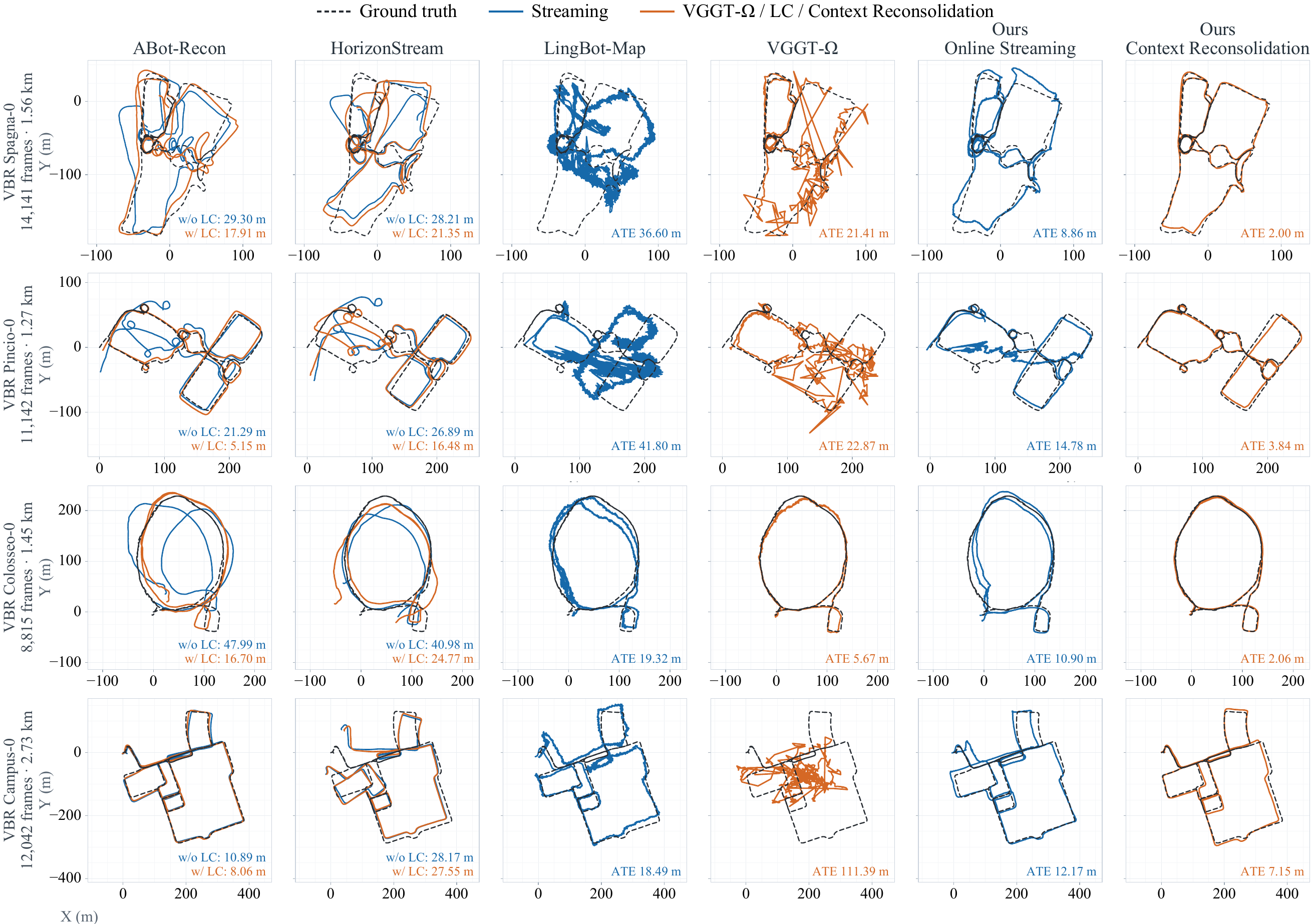}
\captionsetup{skip=3pt,singlelinecheck=false}
\caption{\textbf{Long-trajectory comparisons on four VBR sequences.} Sim(3)-aligned to GT; units in meters.}
\label{fig:trajectory_comparison}
\end{figure}

\subsubsection{Long Trajectory Estimation}
\label{sec:long_trajectory_results}
\label{sec:kitti_vbr_full_results}

We evaluate long-horizon tracking on VBR~\cite{brizi2024vbrvision}, Oxford Spires~\cite{tao2026oxfordspires}, and KITTI~\cite{geiger2012weready}, whose trajectories span urban driving loops, indoor-outdoor campus transitions, and multi-kilometer traversals of up to nearly 19,000 frames. As shown in \cref{tab:kitti_vbr_summary}, even when evaluated on keyframes, offline foundation models degrade substantially over multi-kilometer horizons: trained primarily on bounded view sets, global attention without sequential grounding struggles to preserve consistent scale and orientation across wide-baseline transitions and sparse loops. Among causal streaming approaches, local-window pipelines such as HorizonStream and ABot-Recon maintain smooth short-range motion, yet progressively accumulate drift as older views leave the active window without a global reference. By consolidating incoming keyframes into a unified world frame anchored by $\mathcal A$ and compact trajectory-memory tokens, Ours (Online) curbs long-range drift and outperforms all competing streaming methods on every dataset and metric, nearly halving ATE on VBR from 29.21\,m to 17.01\,m.

Building on this causal state, sequence-end Context Reconsolidation reopens the revisable layers bidirectionally across the retained keyframes in a single feed-forward pass, reducing ATE to 4.72\,m on VBR, 3.92\,m on Oxford Spires, and 9.42\,m on KITTI without place retrieval or pose-graph optimization. It consistently outperforms both full-backbone refinement in SLAM-Former and streaming pipelines paired with external loop-closure backends, including CLoSeR, Scal3R w/ LC, and ABot-Recon w/ LC. The advantage is even more pronounced at the strict AUC@5 threshold: external pose-graph backends redistribute coarse loop residuals across trajectory nodes post hoc without updating internal visual features, leaving residual angular discrepancies across wide-baseline camera pairs. By re-aligning keyframe representations directly in the revisable layers, Ours (Online) already surpasses every loop-closure baseline in AUC@5 on Oxford Spires at 43.41, and Context Reconsolidation more than doubles AUC@5 over the strongest loop-closure competitors on VBR (37.94) and Oxford Spires (74.71), while still leading on KITTI at 41.00. Qualitative comparisons in \cref{fig:trajectory_comparison} and additional trajectories in \cref{fig:supp_trajectory_comparison} illustrate this behavior across VBR, Oxford Spires, and KITTI: whereas offline and sliding-window baselines exhibit severe scale drift or unclosed loops over multi-building paths, Online Streaming preserves global trajectory structure and Context Reconsolidation brings revisited loops into tight alignment with ground truth.

\noindent\textbf{Per-sequence results and extended baselines.}
\Cref{sec:supp_trajectories} (\cref{tab:vbr_comparison,tab:oxford_full_ate,tab:kitti_comparison}) reports per-sequence and mean ATE across all 7 VBR, 10 Oxford Spires, and 11 KITTI trajectories against an extended set of baselines spanning efficient offline foundation models, geometric optimization SLAM systems---several using ground-truth camera intrinsics---and cache-pruning, recurrent, or periodically reset streaming models. Across these extended comparisons, recurrent-state and token-pruning caches progressively lose early geometric context over multi-thousand-frame horizons; periodic 100-frame resets curb state saturation within short segments, yet re-initializing the state breaks global frame continuity across sub-trajectories and leaves a wide gap behind continuous context consolidation in Ours (Online). Compared with multi-stage geometric optimization systems that rely on iterative bundle or pose-graph optimization, Ours (Reconsolidation) achieves substantially lower mean ATE on VBR and KITTI and outperforms all uncalibrated baselines on Oxford Spires in a single feed-forward pass.

\subsubsection{Camera Pose Estimation}
\label{sec:camera_pose_results}

We evaluate scene-level camera pose estimation on CroCoDL~\cite{blum2025crocodlcross}, Tanks and Temples~\cite{knapitsch2017tanksand}, and Princeton365~\cite{kayan2025princeton365diverse} (\cref{sec:scene_pose_protocol,tab:offline_comparison}). On Tanks and Temples and Princeton365, where the full view set fits in GPU memory, offline foundation models provide strong baselines by attending jointly across all views, whereas sliding-window streaming methods lose global reference as the camera orbits around foreground structures. Maintaining long-range context tokens in a unified world frame allows Ours (Online) to preserve cross-orbit registration, nearly doubling AUC@15 over competing streaming baselines on Princeton365 to 41.40 while already achieving a lower 0.15\,m ATE than full-attention offline models. Context Reconsolidation further tightens multi-view alignment across both benchmarks, edging past $\pi^3$ and DA3 on Tanks and Temples at 93.33 AUC@15 and establishing a clear lead across all 57 Princeton365 scanning sequences with 53.17 AUC@15 and 0.09\,m ATE.

\begin{table}[!htb]
\centering
\caption{\textbf{Camera pose estimation.} AUC@15 and AUC@30 (\%) are weighted by valid camera-pair counts; ATE (m) is averaged equally across sequences. Within each group, bold/underline mark best/second best; across all methods, blue/pale blue mark best/second best (ties share color). LC denotes loop closure. ``--'' indicates that LingBot-Map did not report AUC@15 on TAT~\cite{chen2026lingbotmap}.}
\label{tab:offline_comparison}
\label{tab:crocodl_crop}
\small
\setlength{\tabcolsep}{3pt}
\renewcommand{\arraystretch}{1.0}
\begin{adjustbox}{max width=\textwidth}
\begin{tabular}{@{}clccccccccc@{}}
\toprule
 &  & \multicolumn{3}{c}{CroCoDL} & \multicolumn{3}{c}{Tanks and Temples} & \multicolumn{3}{c}{Princeton365} \\
\cmidrule(lr){3-5}\cmidrule(lr){6-8}\cmidrule(lr){9-11}
LC & Method & AUC@15$\uparrow$ & AUC@30$\uparrow$ & ATE$\downarrow$ & AUC@15$\uparrow$ & AUC@30$\uparrow$ & ATE$\downarrow$ & AUC@15$\uparrow$ & AUC@30$\uparrow$ & ATE$\downarrow$ \\
\midrule
\multicolumn{11}{@{}l}{\textit{Offline}} \\
\ding{55} & VGGT-$\Omega$ & \cellcolor[HTML]{E8F1FA}\textbf{58.08} & \cellcolor[HTML]{E8F1FA}\textbf{70.18} & \textbf{1.26} & 82.13 & 84.20 & 0.65 & \underline{48.32} & \underline{56.51} & \textbf{0.17} \\
\ding{55} & $\pi^{3}$ & \underline{41.83} & \underline{54.67} & \underline{2.11} & \underline{91.86} & \cellcolor[HTML]{E8F1FA}\textbf{95.08} & \cellcolor[HTML]{E8F1FA}\underline{0.15} & 44.20 & 52.58 & \underline{0.22} \\
\ding{55} & DA3 (Nested 1.1) & 40.62 & 52.90 & 2.83 & \cellcolor[HTML]{E8F1FA}\textbf{92.72} & \underline{95.01} & \cellcolor[HTML]{BFD9F2}\textbf{0.14} & \cellcolor[HTML]{E8F1FA}\textbf{48.85} & \textbf{57.35} & \textbf{0.17} \\
\midrule
\multicolumn{11}{@{}l}{\textit{Streaming}} \\
\ding{55} & SLAM-Former (frontend) & 6.95 & 15.39 & 5.95 & \underline{50.03} & 60.77 & 0.83 & 15.99 & 28.04 & 0.33 \\
\ding{55} & LingBot-Map & \underline{34.42} & \underline{49.05} & \underline{2.12} & -- & \underline{92.34} & \underline{0.21} & 22.03 & 34.77 & \underline{0.20} \\
\ding{55} & HorizonStream & 11.44 & 20.92 & 4.48 & 49.66 & 68.16 & 0.67 & 15.88 & 29.84 & 0.28 \\
\ding{55} & ABot-Recon & 17.91 & 32.49 & 3.41 & 49.50 & 69.24 & 0.55 & \underline{22.38} & \underline{42.16} & 0.21 \\
\ding{55} & \textbf{Ours (Online)} & \textbf{50.70} & \textbf{66.82} & \cellcolor[HTML]{E8F1FA}\textbf{1.11} & \textbf{87.67} & \textbf{92.99} & \textbf{0.19} & \textbf{41.40} & \textbf{53.09} & \cellcolor[HTML]{E8F1FA}\textbf{0.15} \\
\midrule
\multicolumn{11}{@{}l}{\textit{Global refinement}} \\
\ding{55} & SLAM-Former (w/ backend) & 15.13 & 26.39 & 4.45 & \underline{84.61} & \underline{91.22} & \underline{0.17} & 36.62 & 47.17 & 0.25 \\
\ding{51} & CLoSeR & 15.72 & 27.68 & 4.42 & 63.09 & 77.25 & 0.45 & \underline{41.71} & \cellcolor[HTML]{E8F1FA}\underline{61.75} & \underline{0.16} \\
\ding{51} & HorizonStream w/ LC & 9.47 & 18.57 & 4.37 & 48.12 & 67.84 & 0.75 & 10.18 & 20.29 & 0.31 \\
\ding{51} & ABot-Recon w/ LC & \underline{17.92} & \underline{32.51} & \underline{3.41} & 49.72 & 69.47 & 0.55 & 22.57 & 42.67 & 0.21 \\
\ding{55} & \textbf{Ours (Reconsolidation)} & \cellcolor[HTML]{BFD9F2}\textbf{74.09} & \cellcolor[HTML]{BFD9F2}\textbf{85.30} & \cellcolor[HTML]{BFD9F2}\textbf{0.44} & \cellcolor[HTML]{BFD9F2}\textbf{93.33} & \cellcolor[HTML]{BFD9F2}\textbf{95.97} & \cellcolor[HTML]{BFD9F2}\textbf{0.14} & \cellcolor[HTML]{BFD9F2}\textbf{53.17} & \cellcolor[HTML]{BFD9F2}\textbf{67.10} & \cellcolor[HTML]{BFD9F2}\textbf{0.09} \\
\bottomrule
\end{tabular}
\end{adjustbox}
\end{table}

The gain from context reconsolidation is especially pronounced on CroCoDL, whose handheld sequences traverse dimly lit basements, semi-collapsed ruins, and open plazas under harsh lighting transitions and rapid viewpoint changes. In these cluttered trajectories, external loop-closure backends relying on visual place retrieval rarely establish reliable loop constraints, leaving ABot-Recon w/ LC and HorizonStream w/ LC nearly unchanged from their streaming frontends. Even in causal mode, Ours (Online) reaches 1.11\,m ATE, outperforming both retrieval-based refinement pipelines and offline VGGT-$\Omega$. Bypassing retrieval gates, Context Reconsolidation attends directly across all retained keyframes in the revisable layers, boosting AUC@15 from 50.70 to 74.09 and AUC@30 from 66.82 to 85.30, cutting ATE to 0.44\,m, and surpassing VGGT-$\Omega$ by over 15 AUC@30 points.

\subsection{Dense Reconstruction}
\label{sec:small_scene_results}

We evaluate dense 3D reconstruction on CroCoDL, EuRoC Vicon~\cite{burri2016eurocmicro}, and Oxford Spires (\cref{sec:dense_protocol,tab:oxford_reconstruction}), spanning unstructured indoor--outdoor ruins, agile room-scale micro-aerial flights with millimeter-accurate laser ground truth, and multi-building campus traversals. Because unprojecting per-view depths into a shared 3D space couples local depth precision with cross-view pose alignment, accumulated drift in sliding-window baselines projects repeated structures to offset locations, producing duplicated walls and coverage gaps. Consolidating keyframes in a unified world frame online allows Ours (Online) to reach 67.03 F1 on CroCoDL and 94.63 F1 on Oxford Spires---already surpassing both streaming and offline baselines on these extended trajectories---while matching the best 0.03\,m completeness on EuRoC.

\begin{table}[!htb]
\centering
\caption{\textbf{Dense reconstruction.} Sequence-averaged accuracy (Acc, m), completeness (Comp, m), and F1 (\%). F1 thresholds are 0.25\,m for CroCoDL and EuRoC and 4\,m for Oxford. Within each group, bold/underline mark best/second best; across all methods, blue/pale blue mark best/second best (ties share color).}
\label{tab:oxford_reconstruction}
\small
\setlength{\tabcolsep}{3pt}
\renewcommand{\arraystretch}{1.0}
\begin{tabular*}{\textwidth}{@{\extracolsep{\fill}}lccccccccc@{}}
\toprule
 & \multicolumn{3}{c}{CroCoDL} & \multicolumn{3}{c}{EuRoC} & \multicolumn{3}{c}{Oxford} \\
\cmidrule(lr){2-4}\cmidrule(lr){5-7}\cmidrule(lr){8-10}
\hspace*{1em}Method & Acc$\downarrow$ & Comp$\downarrow$ & F1$\uparrow$ & Acc$\downarrow$ & Comp$\downarrow$ & F1$\uparrow$ & Acc$\downarrow$ & Comp$\downarrow$ & F1$\uparrow$ \\
\midrule
\multicolumn{10}{@{}l}{\textit{Streaming}} \\
\hspace*{1em}SLAM-Former (frontend) & 0.61 & 1.13 & 39.53 & 0.20 & \underline{0.05} & 81.15 & 3.72 & 4.56 & 68.50 \\
\hspace*{1em}LingBot-Map & \underline{0.44} & \underline{0.66} & \underline{52.95} & \textbf{0.16} & \cellcolor[HTML]{BFD9F2}\textbf{0.03} & \textbf{88.02} & 1.78 & \underline{0.95} & \underline{92.30} \\
\hspace*{1em}HorizonStream & 0.55 & 1.65 & 32.20 & 0.25 & 0.07 & 72.02 & \underline{1.58} & 1.32 & 91.70 \\
\hspace*{1em}ABot-Recon & 0.63 & 1.18 & 37.28 & 0.24 & \underline{0.05} & 74.61 & 1.67 & 0.98 & 91.43 \\
\hspace*{1em}\textbf{Ours (Online)} & \cellcolor[HTML]{E8F1FA}\textbf{0.38} & \cellcolor[HTML]{E8F1FA}\textbf{0.34} & \cellcolor[HTML]{E8F1FA}\textbf{67.03} & \underline{0.18} & \cellcolor[HTML]{BFD9F2}\textbf{0.03} & \underline{85.72} & \cellcolor[HTML]{E8F1FA}\textbf{1.35} & \cellcolor[HTML]{BFD9F2}\textbf{0.65} & \cellcolor[HTML]{E8F1FA}\textbf{94.63} \\
\midrule
\multicolumn{10}{@{}l}{\textit{Offline and reconsolidation}} \\
\hspace*{1em}VGGT-$\Omega$ & \underline{0.43} & \underline{0.45} & \underline{60.69} & \cellcolor[HTML]{E8F1FA}\underline{0.10} & \underline{0.05} & 94.96 & \underline{2.24} & 3.03 & 84.47 \\
\hspace*{1em}$\pi^3$ & 0.45 & 0.60 & 56.25 & \cellcolor[HTML]{BFD9F2}\textbf{0.09} & \cellcolor[HTML]{E8F1FA}\textbf{0.04} & \cellcolor[HTML]{E8F1FA}\underline{96.63} & 2.32 & 1.87 & 83.77 \\
\hspace*{1em}DA3 & 0.45 & 0.60 & 59.33 & 0.13 & \cellcolor[HTML]{E8F1FA}\textbf{0.04} & 91.69 & 2.29 & \underline{1.36} & \underline{87.70} \\
\hspace*{1em}\textbf{Ours (Reconsolidation)} & \cellcolor[HTML]{BFD9F2}\textbf{0.31} & \cellcolor[HTML]{BFD9F2}\textbf{0.27} & \cellcolor[HTML]{BFD9F2}\textbf{77.14} & \cellcolor[HTML]{BFD9F2}\textbf{0.09} & \cellcolor[HTML]{E8F1FA}\textbf{0.04} & \cellcolor[HTML]{BFD9F2}\textbf{97.71} & \cellcolor[HTML]{BFD9F2}\textbf{1.03} & \cellcolor[HTML]{E8F1FA}\textbf{0.66} & \cellcolor[HTML]{BFD9F2}\textbf{95.46} \\
\bottomrule
\end{tabular*}
\end{table}

Context Reconsolidation re-applies the shared multi-scale DPT head to fuse the invariant committed features from lower layers, which preserve sharp per-view local geometry without re-encoding images, with the bidirectionally updated revisable features from upper layers, which resolve global pose and scale drift. Examining Accuracy and Completeness separately clarifies how reconsolidation improves surface geometry across scales. On room-scale EuRoC, short indoor trajectories already let online streaming cover the room densely, and reconsolidation halves Acc from 0.18\,m to 0.09\,m to reach 97.71 F1, edging past $\pi^3$ and VGGT-$\Omega$. On campus-scale Oxford Spires, offline models suffer degraded completeness over multi-thousand-frame horizons, whereas Ours (Reconsolidation) reduces Acc to 1.03\,m while preserving 0.66\,m completeness at 95.46 F1. On CroCoDL, resolving multi-view drift across cluttered, poorly lit scenes lowers Acc to 0.31\,m and Comp to 0.27\,m, reaching 77.14 F1 and surpassing VGGT-$\Omega$ by over 16 points.

\noindent\textbf{Qualitative comparisons.}
\label{sec:more_qualitative_3d}
\Cref{fig:more_3d_reconstruction} and \cref{fig:more_3d_reconstruction_extra} (\cref{sec:supp_qualitative}) compare dense point clouds and camera trajectories from \method, ABot-Recon, and VGGT-$\Omega$ across 12 outdoor, mixed indoor--outdoor, and indoor sequences spanning urban driving loops, campus courtyards and arcades, narrow street corridors, gardens, and multi-room interiors. Across multi-block loops and indoor--outdoor transitions, ABot-Recon accumulates drift once early keyframes leave its sliding window, leaving revisited streets unclosed and splitting repeated facades into double surfaces, while VGGT-$\Omega$ exhibits global scale warping and fragmented sub-trajectories over long horizons. In contrast, \method closes large-scale loops cleanly without pose-graph optimization, preserving planar ground surfaces, crisp architectural facades, and coherent 3D structure across scales.

\endgroup

\begin{figure}[p]
\centering
\IfFileExists{fig/qualitative-3d-reconstruction.pdf}{%
  \includegraphics[width=\textwidth]{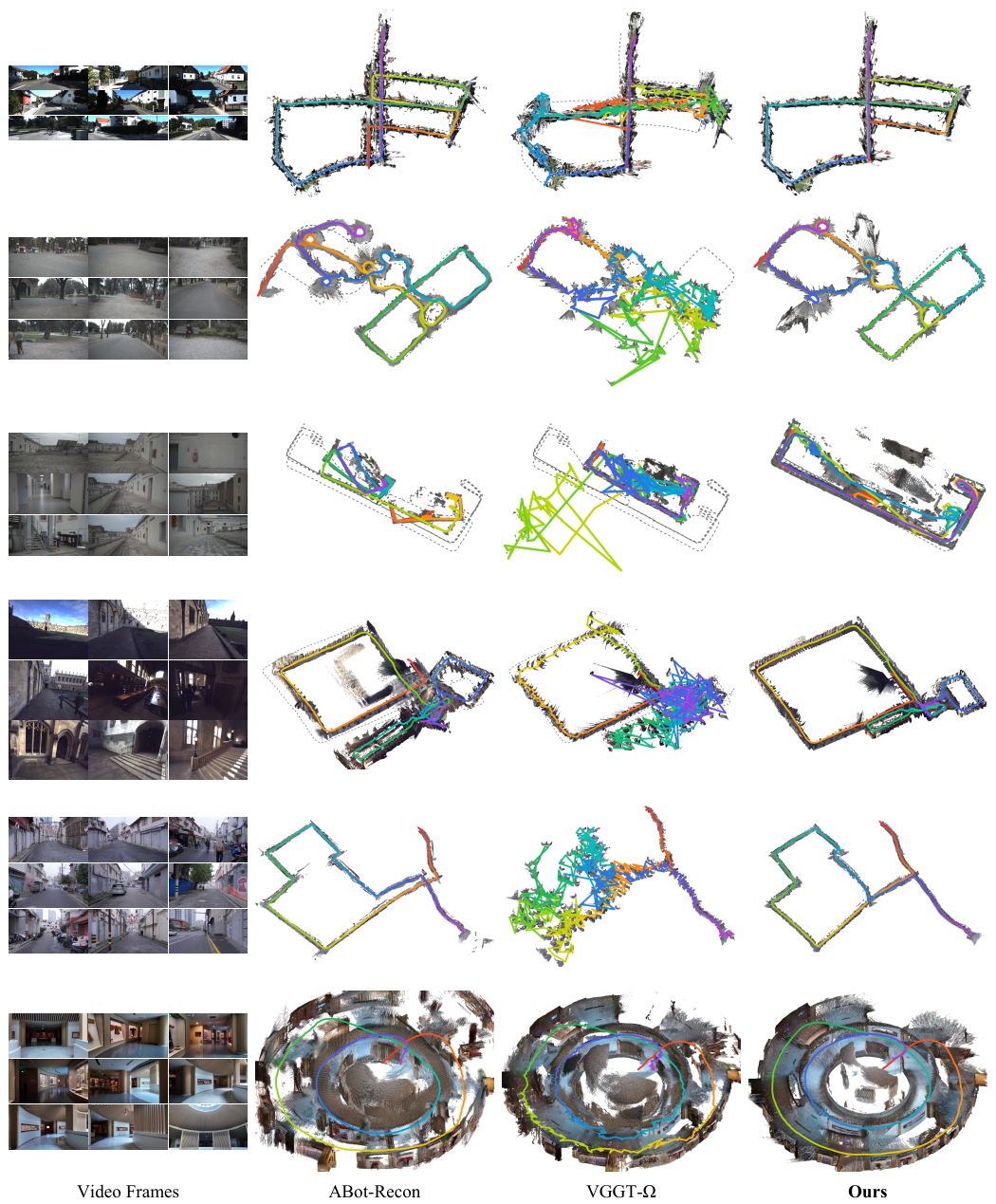}%
}{%
  \fbox{\parbox[c][0.30\textheight][c]{0.96\textwidth}{\centering\placeholder{Upload multi-scene 3D reconstruction comparison to \texttt{fig/qualitative-3d-reconstruction.pdf}}}}%
}
\caption{\textbf{Qualitative 3D reconstruction comparisons.} Sampled input frames (left) and reconstructed 3D point clouds overlaid with camera trajectories for ABot-Recon, VGGT-$\Omega$, and \method across large-scale outdoor driving/campus sequences and a multi-loop indoor scene.}
\label{fig:more_3d_reconstruction}
\end{figure}

\subsection{Ablation Studies}
\label{sec:ablations}

We conduct ablation studies to examine our training design, mid-stream context write-back, the layer split between committed and revisable stages, and visual encoder initialization (\cref{tab:ablation_components,tab:midstream_panel,tab:bidirectional_depth,tab:pretraining_initialization}).

\noindent\textbf{Training components.}
Using dedicated checkpoints trained under a lighter schedule---distinct from our full-schedule main model---we start from LingBot-Map~\cite{chen2026lingbotmap} as the causal baseline and evaluate progressive training choices on KITTI in \cref{tab:ablation_components}. Applying bidirectional reconsolidation directly to the purely causal baseline weights causes trajectory accuracy to collapse, with ATE rising from 24.04\,m to 72.28\,m: upper layers trained solely under causal masks suffer severe distribution mismatch when exposed to future keys and values at inference time. Extending causal training to longer sequences substantially curbs online drift, and adding dense adjacent relative-pose supervision in \cref{eq:relative_loss} cuts frame-to-frame translational jitter RPE-t nearly in half from 1.38\,m to 0.81\,m by constraining consecutive motion increments alongside global scale-normalized poses. Even so, causal streaming training alone leaves zero-shot reconsolidation degraded. Incorporating mixed-mode continuous training (\cref{sec:training}) exposes the revisable layers to causal, bidirectional, and reconsolidated-prefix attention patterns within a single set of weights, reducing GCR ATE to 14.96\,m and RPE-t to 0.20\,m while also achieving the lowest online trajectory error.

\begin{table}[!htbp]
\centering
\caption{\textbf{Cumulative training components on KITTI.} Long: long-sequence training; Adj.: adjacent-pair supervision; Mix: mixed-mode training. Components are added progressively from the baseline (first row) under the lighter training schedule. ATE and RPE-t are in meters; AUC@30 is in percent. Bold/underline mark best/second best.}
\label{tab:ablation_components}
\label{tab:ablation_overview}
\small
\setlength{\tabcolsep}{4pt}
\renewcommand{\arraystretch}{1.0}
\begin{tabular*}{\linewidth}{@{\extracolsep{\fill}}cccrrrrrr@{}}
\toprule
\multicolumn{3}{c}{Training components} & \multicolumn{3}{c}{Online Streaming} & \multicolumn{3}{c}{Context Reconsolidation} \\
\cmidrule(lr){1-3}\cmidrule(lr){4-6}\cmidrule(l){7-9}
Long & Adj. & Mix & ATE$\downarrow$ & AUC@30$\uparrow$ & RPE-t$\downarrow$ & ATE$\downarrow$ & AUC@30$\uparrow$ & RPE-t$\downarrow$ \\
\midrule
-- & -- & -- & 24.04 & 73.56 & 2.84 & 72.28 & 44.58 & 0.98 \\
\ding{51} & -- & -- & 16.50 & \textbf{78.57} & 1.38 & \underline{62.02} & \underline{48.52} & 0.57 \\
\ding{51} & \ding{51} & -- & \underline{16.42} & 76.52 & \underline{0.81} & 63.62 & 47.28 & \underline{0.49} \\
\ding{51} & \ding{51} & \ding{51} & \textbf{15.49} & \underline{78.00} & \textbf{0.76} & \textbf{14.96} & \textbf{78.53} & \textbf{0.20} \\
\bottomrule
\end{tabular*}
\end{table}

\noindent\textbf{Streaming after Reconsolidation.}
\Cref{tab:midstream_panel} evaluates whether writing the revised context $\hat{\mathcal G}^{\mathrm r}_\tau$ back into the active cache benefits subsequent views across all 7 VBR, 10 Oxford Spires, and 11 KITTI sequences. Each trajectory is split at 75\% of its retained keyframes under one full-trajectory Sim(3) alignment per mode. Streaming after GCR triggers one reconsolidation at the split and then resumes causal inference without further revision. Continuation ATE decreases from 23.58\,m to 14.87\,m on VBR, from 6.96\,m to 4.33\,m on Oxford Spires, and from 21.21\,m to 14.22\,m on KITTI. On VBR, history ATE also falls from 13.57\,m to 6.63\,m. These continuation frames are inferred causally after the trigger, so the comparison evaluates the effect of conditioning future predictions on the rewritten context. Full GCR provides a sequence-end reference that also uses observations from the final quarter.

\begin{table}[!htbp]
\centering
\caption{\textbf{Streaming after Context Reconsolidation across long-trajectory benchmarks.} History and continuation ATE (m, $\downarrow$), averaged equally over 7 VBR, 10 Oxford Spires, and 11 KITTI sequences. The split is at 75\% of retained keyframes, with one full-trajectory Sim(3) alignment per sequence and mode. Streaming after GCR revises once at the split and resumes online; Full GCR revises at sequence end. Bold marks the lowest ATE in each column.}
\label{tab:midstream_panel}
\small
\setlength{\tabcolsep}{3pt}
\renewcommand{\arraystretch}{1.0}
\begin{tabular*}{\linewidth}{@{\extracolsep{\fill}}lrrrrrr@{}}
\toprule
& \multicolumn{2}{c}{VBR} & \multicolumn{2}{c}{Oxford Spires} & \multicolumn{2}{c}{KITTI} \\
\cmidrule(lr){2-3}\cmidrule(lr){4-5}\cmidrule(l){6-7}
Mode & History & Continuation & History & Continuation & History & Continuation \\
\midrule
Online & 13.57 & 23.58 & 5.52 & 6.96 & 15.77 & 21.21 \\
Streaming after GCR & 6.63 & 14.87 & 4.56 & 4.33 & 9.89 & 14.22 \\
Full GCR & \textbf{4.98} & \textbf{3.66} & \textbf{4.21} & \textbf{2.85} & \textbf{9.46} & \textbf{8.40} \\
\bottomrule
\end{tabular*}
\end{table}

\noindent\textbf{Number of revisable layers.}
\Cref{tab:bidirectional_depth} studies where to partition the 24-layer backbone into committed and revisable stages on KITTI under the same lighter schedule. Leaving only 8 revisable layers restricts cross-view alignment capacity and increases trajectory error to 17.40\,m ATE. Conversely, reopening all 24 layers slightly improves per-keyframe depth AbsRel from 0.077 to 0.074, yet degrades trajectory ATE from 14.96\,m to 16.06\,m while doubling reconsolidation compute and forfeiting committed-cache invariance. This trade-off reflects the specialization of layers across depth: lower causal layers already extract reliable local geometry across smoothly varying video streams and provide a stable coordinate anchor, whereas upper layers govern global cross-view registration where drift accumulates. Splitting the backbone evenly into 12 committed and 12 revisable layers achieves the lowest trajectory error while reusing half of the backbone cache without recomputation.\par

\noindent\textbf{Visual encoder initialization.}
\label{sec:pretraining_initialization}
We compare DINOv3-L~\cite{simeoni2025dinov3} and LingBot-Vision-L~\cite{fu2026vision} as visual encoder initializations for multi-view offline training under an identical 16k-iteration schedule with a longer side of 512 and patch size 16. Across 7-Scenes, NRGBD, ETH3D, HiRoom, and Sintel, we report single-scale-aligned per-sequence AbsRel and pose AUC over all valid within-sequence camera pairs in \cref{tab:pretraining_initialization}. LingBot-Vision-L improves depth accuracy and yields a clear 5.02-point gain at the strict AUC@3 threshold, rising from 50.77 to 55.79. Because small angular errors in local multi-view perception compound over thousand-frame streams, this stronger fine-grained geometric prior provides a better starting point for subsequent streaming and mixed-mode continuous training.

\begin{table}[!htbp]
\centering
\begin{minipage}[t]{.47\textwidth}
\vspace{0pt}
\centering
\caption{\textbf{Committed and revisable layer splits on KITTI.} The 24-layer backbone is evaluated under the lighter training schedule. Reconsolidated trajectory ATE is in meters. Lower is better.}
\label{tab:bidirectional_depth}
\small
\setlength{\tabcolsep}{5pt}
\renewcommand{\arraystretch}{1.0}
\begin{tabular}{@{}ccrr@{}}
\toprule
\multicolumn{2}{c}{Layers} & \multicolumn{2}{c}{GCR} \\
\cmidrule(r){1-2}\cmidrule(l){3-4}
Committed & Revisable & ATE & AbsRel \\
\midrule
16 & 8 & 17.40 & 0.077 \\
12 & 12 & \textbf{14.96} & 0.077 \\
0 & 24 & 16.06 & \textbf{0.074} \\
\bottomrule
\end{tabular}
\end{minipage}\hfill
\begin{minipage}[t]{.50\textwidth}
\vspace{0pt}
\centering
\caption{\textbf{Visual initialization for multi-view offline training.} AbsRel and pose AUC (\%) are averaged over five equally weighted datasets: 7-Scenes~\cite{shotton2013scenecoordinate}, NRGBD~\cite{azinovic2022neuralrgbd}, ETH3D~\cite{schops2017multiview}, HiRoom~\cite{lin2026depthanything}, and Sintel~\cite{butler2012naturalistic}.}
\label{tab:pretraining_initialization}
\small
\setlength{\tabcolsep}{3pt}
\renewcommand{\arraystretch}{1.0}
\begin{tabular}{@{}lrrr@{}}
\toprule
Initialization & AbsRel$\downarrow$ & AUC@3$\uparrow$ & AUC@30$\uparrow$ \\
\midrule
DINOv3 & 0.0732 & 50.77 & 85.72 \\
LingBot-Vision & \textbf{0.0726} & \textbf{55.79} & \textbf{86.55} \\
\bottomrule
\end{tabular}
\end{minipage}
\end{table}

\FloatBarrier

\subsection{Revisit Consistency Evaluation}
\label{sec:revisit_evaluation}

Whole-trajectory Sim(3) ATE distributes error across an entire sequence and can therefore mask local discontinuities where a camera returns to a previously visited location. We therefore evaluate local pose consistency directly across the ground-truth revisit pairs defined in \cref{sec:evaluation_details}.

\begin{table}[!htbp]
\centering
\caption{\textbf{Pose accuracy at shared GT-defined revisits.} ATE and $e_t$ are in meters, $e_R$ in degrees. VBR uses all seven sequences; KITTI uses the six sequences containing revisits (\texttt{00}, \texttt{02}, \texttt{05}, \texttt{06}, \texttt{07}, and \texttt{09}).}
\label{tab:revisit_cases}
\small
\setlength{\tabcolsep}{5pt}
\renewcommand{\arraystretch}{1.0}
\begin{tabular*}{\linewidth}{@{\extracolsep{\fill}}l*{6}{c}@{}}
\toprule
 & \multicolumn{3}{c}{VBR} & \multicolumn{3}{c}{KITTI} \\
\cmidrule(lr){2-4}\cmidrule(l){5-7}
Method & ATE$\downarrow$ & $e_t\!\downarrow$ & $e_R\!\downarrow$
 & ATE$\downarrow$ & $e_t\!\downarrow$ & $e_R\!\downarrow$ \\
\midrule
HorizonStream & 29.21 & 32.81 & 28.87 & 30.36 & 72.63 & 17.33 \\
ABot-Recon & 29.89 & 36.45 & 24.07 & 22.34 & 34.50 & 8.85 \\
\textbf{Ours (Online)} & \textbf{17.01} & \textbf{14.56} & \textbf{11.76} & \textbf{16.98} & \textbf{24.29} & \textbf{1.77} \\
\midrule
HorizonStream w/ LC & 17.78 & 16.31 & 22.46 & 22.08 & 15.72 & 13.68 \\
ABot-Recon w/ LC & 9.98 & 7.02 & 12.14 & 13.61 & 7.42 & 5.76 \\
\textbf{Ours (Reconsolidation)} & \textbf{4.72} & \textbf{1.56} & \textbf{5.47} & \textbf{7.26} & \textbf{5.66} & \textbf{0.93} \\
\bottomrule
\end{tabular*}
\end{table}

\noindent\textbf{Loop-closure accuracy.}
Under full-trajectory Sim(3) alignment, \cref{tab:revisit_cases} reports whole-trajectory ATE alongside the local translational error $e_t$ and rotational error $e_R$ of the relative pose $T_i^{-1}T_j$ across matched revisit pairs on VBR and KITTI. Even in causal Online mode, consolidating observations in a shared world frame yields far smaller revisit discrepancies than sliding-window baselines, whose window-to-window stitching accumulates large offsets before the camera returns; notably, Ours (Online) already keeps revisit rotation error to $1.77^\circ$ on KITTI, beating both competing pipelines even with loop closure enabled. Optional retrieval and pose-graph stages in HorizonStream w/ LC and ABot-Recon w/ LC reduce drift, yet local misalignment persists at revisit junctions. Without place recognition or pose-graph optimization, Context Reconsolidation links distant views directly in the revisable layers, cutting revisit error over fourfold vs.\ ABot-Recon w/ LC on VBR to 1.56\,m $e_t$ and reaching $0.93^\circ$ $e_R$ on KITTI.\par

\label{sec:revisit_visual_details}
\begin{figure}[!htb]
\centering
\includegraphics[width=\linewidth]{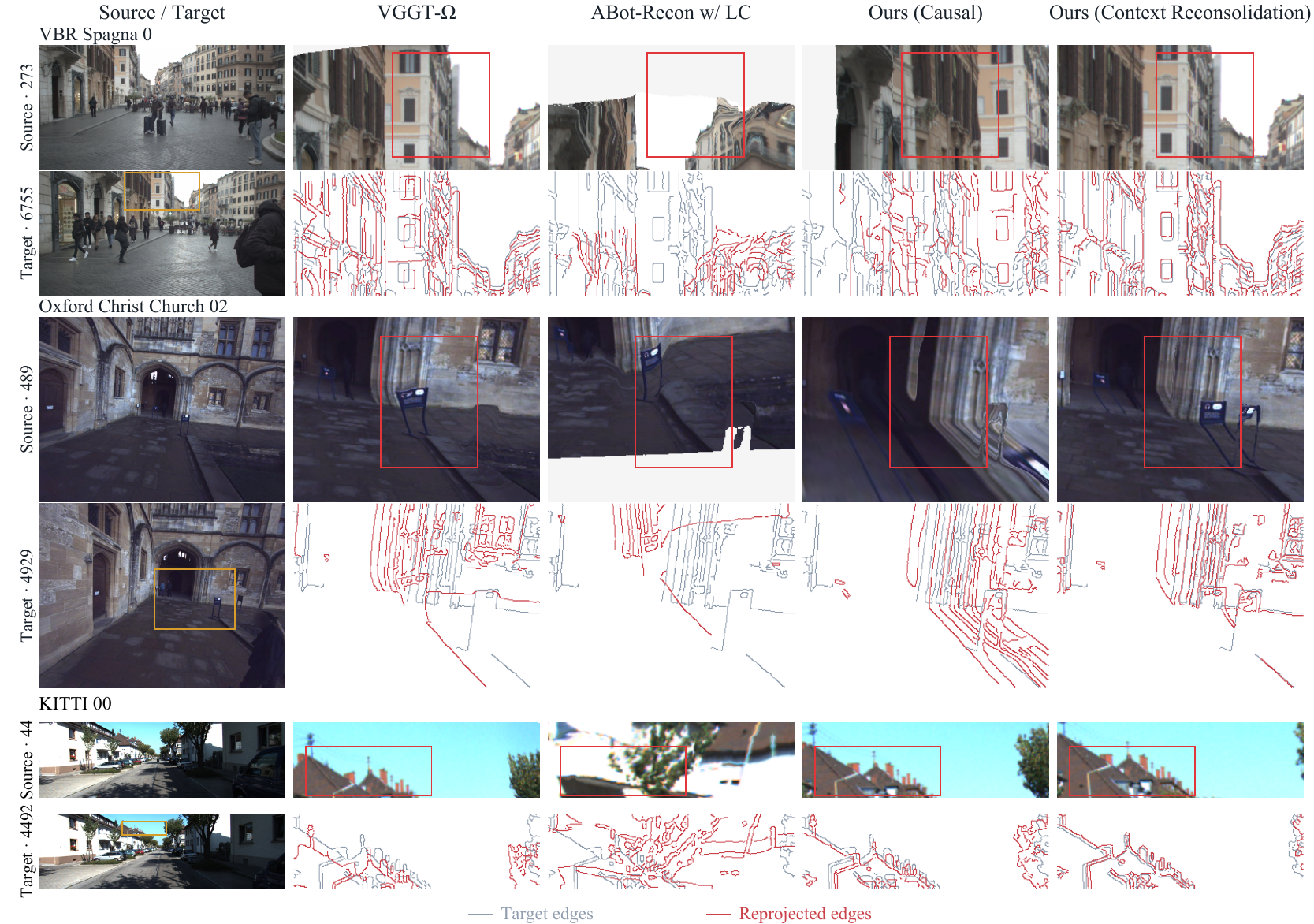}
\caption{\textbf{Alignment across revisits on VBR, Oxford, and KITTI.} Reprojections of source images into revisited target views. Yellow boxes locate the crops; red boxes highlight corresponding structures. Gray and red edges denote target and reprojected contours.}
\label{fig:cross_view_alignment}
\end{figure}

\noindent\textbf{Alignment across revisits.}
\Cref{fig:cross_view_alignment} reprojects source views into revisited target views on VBR Spagna-0 (273 to 6,755), Oxford Christ Church-02 (489 to 4,929), and KITTI 00 (44 to 4,492), using predicted depths and relative poses directly. Because 3D reprojection couples per-view depth scale with long-range relative pose accuracy, any drift in translation, rotation, or scale appears immediately as shifted contours. Context Reconsolidation brings archways, rooflines, and window frames into clean register compared with causal streaming, whereas VGGT-$\Omega$ and ABot-Recon w/ LC retain noticeable double edges and structural offsets.\par

\begin{figure}[!htb]
\centering
\includegraphics[width=\linewidth]{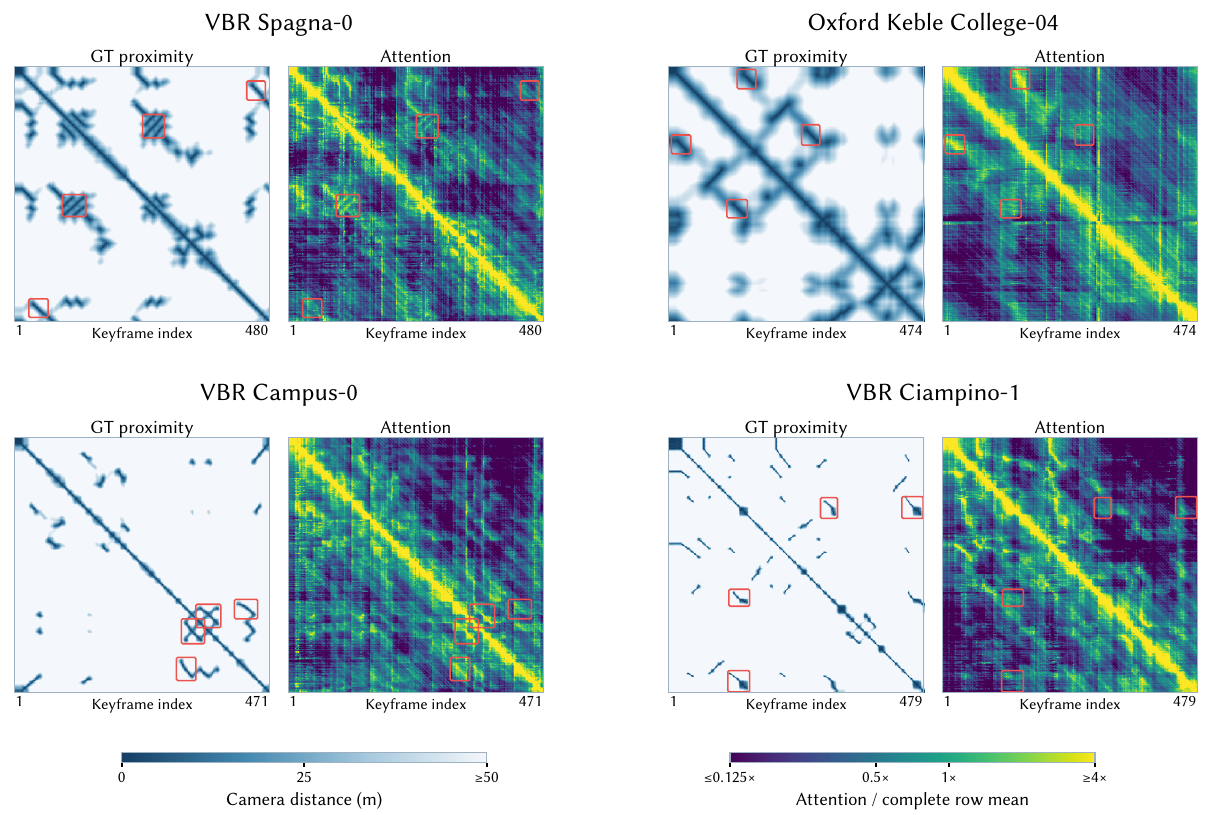}
\caption{\textbf{Attention across revisits on VBR and Oxford.} Full-sequence GT proximity and layer-16 bidirectional attention from the same checkpoint and keyframe sets as the main evaluation; corresponding GT-defined revisit regions are boxed in red. Attention is divided by the complete row mean and displayed on a shared log scale clipped to $[0.125,4]$.}
\label{fig:revisit_attention}
\end{figure}

\noindent\textbf{Emergent revisit attention.}
\Cref{fig:revisit_attention} compares ground-truth camera proximity with head-averaged layer-16 bidirectional GCR attention weights on VBR Spagna-0, Campus-0, Ciampino-1, and Oxford Keble College-04. All four scenes use the main evaluation checkpoint, input preprocessing, and retained keyframes, and the extracted run reproduces the archived reconsolidated poses. Even without explicit loop-closure or place-retrieval supervision, sharp off-diagonal bands emerge wherever the camera returns to spatially proximate locations thousands of frames later. Mirroring ground-truth proximity across multi-pass loops, these bands show that the revisable layers associate co-visible geometry directly from token representations to resolve accumulated drift.

\section{Conclusion and Discussion}
\label{sec:conclusion}

We have presented \method, a feed-forward foundation model for 3D reconstruction from continuous visual streams. At its core is Geometric Context Reconsolidation (GCR), which resolves accumulated streaming drift by selectively reconsolidating the cached geometric context that produced past estimates rather than post-processing output trajectories externally. Partitioning the backbone into an invariant lower \emph{committed context} and a selectively reopened upper \emph{revisable context} allows \method to reuse causal per-view geometry without re-encoding images or rerunning lower layers, while recomputing cross-view alignment bidirectionally in a single feed-forward pass and writing the updated state back into the active cache. Trained with our geometric streaming base and mixed-mode continuous training recipe, alongside sequence scale normalization and dense adjacent relative-pose supervision, a single set of weights supports Online Streaming, Context Reconsolidation, and Streaming after Reconsolidation, delivering real-time online inference alongside state-of-the-art long-trajectory tracking, scene-level pose estimation, detector-free loop closure, and dense 3D reconstruction.

\noindent\textbf{Limitations and Future Work.}
While \method performs strongly across diverse indoor and outdoor benchmarks, several limitations remain. First, although image-patch pages outside the local window are recycled in constant space, the retained keyframe context tokens and archived committed features $H_k$ still grow linearly with the number of consolidated keyframes, and full bidirectional attention during reconsolidation scales quadratically with $|\mathcal K_\tau|$. Combining GCR with bounded or hierarchical memory management, such as merging redundant revisits or restricting reconsolidation to topologically relevant sub-contexts, is a promising path toward lifelong operation on unbounded streams. Second, \method currently operates solely on uncalibrated monocular RGB streams, recovering geometry up to an unknown global scale and relying purely on visual cues under rapid motion or low texture; integrating multi-modal inputs such as IMU or range measurements to anchor metric scale and motion priors, and coupling the reconsolidated spatial context with downstream embodied navigation, are natural directions for future work.

\FloatBarrier

{\small
\bibliographystyle{plain}
\bibliography{ref.bib}
}

\appendix
\setcounter{table}{0}
\renewcommand{\thetable}{S\arabic{table}}
\renewcommand{\theHtable}{supp.\arabic{table}}
\setcounter{figure}{0}
\renewcommand{\thefigure}{S\arabic{figure}}
\renewcommand{\theHfigure}{supp.\arabic{figure}}
\setcounter{equation}{0}
\renewcommand{\theequation}{S\arabic{equation}}
\renewcommand{\theHequation}{supp.\arabic{equation}}
\crefalias{section}{appendix}

\section{Additional Trajectory Results}
\label{sec:supp_trajectories}

\Cref{tab:vbr_comparison,tab:oxford_full_ate,tab:kitti_comparison} report per-sequence and mean Sim(3)-aligned ATE across all 7 VBR~\cite{brizi2024vbrvision}, 10 Oxford Spires~\cite{tao2026oxfordspires}, and 11 KITTI~\cite{geiger2012weready} trajectories, expanding the summary comparison in \cref{sec:long_trajectory_results} with additional baselines across three method families. For offline foundation models, we include VGGT~\cite{wang2025vggtvisual}, accelerated variants VGG-T$^{3}$~\cite{elflein2026vggt} and FastVGGT~\cite{shen2026fastvggtfast}, and the updated $\pi^{3}$ release $\pi^{3}$X~\cite{wang20263permutation}, all evaluated on the same sampled keyframes with released weights, native preprocessing, and linear camera-center plus SLERP~\cite{shoemake1985animatingrotation} interpolation. For geometric optimization systems, we evaluate submap and SLAM pipelines built on 3D foundation models---VGGT-SLAM~\cite{maggio2025vggtslam}, MASt3R-SLAM~\cite{murai2025mast3rslam}, VGGT-Long~\cite{deng2026vggtlong}, and SURE-Map~\cite{liu2026suremap}---alongside classical deep visual odometry and SLAM systems that utilize known ground-truth camera intrinsics ($^{*}$), namely DROID-SLAM~\cite{teed2021droidslam}, DPV-SLAM~\cite{lipson2024deeppatch}, DPVO~\cite{teed2023deeppatch}, and DROID-W~\cite{li2026droidslam}. For causal streaming models, we additionally compare token-pruning and windowed-attention pipelines InfiniteVGGT~\cite{yuan2026infinitevggtvisual}, OVGGT~\cite{lu2026ovggto}, and Stream3R-w~\cite{lan2026stream3rscalable}, recurrent-state models CUT3R~\cite{wang2025continuous3d} and TTT3R~\cite{chen2026ttt3r3d} together with their periodic 100-frame reset variants, and scale-stabilized LongStream~\cite{cheng2026longstreamlong}.

Examining per-sequence accuracy clarifies how memory design governs long-horizon stability. While offline models and recurrent or token-pruning streaming baselines achieve reasonable accuracy on short sub-kilometer trajectories such as KITTI~\texttt{03} and \texttt{04}, their error grows sharply on multi-kilometer sequences spanning thousands of frames, such as VBR Ciampino-1, Oxford Keble College-04 and Christ Church-02, and KITTI~\texttt{00}, \texttt{02}, and \texttt{08}. Periodic 100-frame resets in CUT3R and TTT3R alleviate recurrent state saturation locally, reducing mean ATE on VBR from nearly 100\,m to 70.40\,m and 65.50\,m, yet re-initializing the hidden state severs global coordinate continuity across segments. By maintaining an anchored world coordinate frame with paged local patch windows and compact keyframe context tokens, Ours (Online) sustains coherent tracking over multi-kilometer horizons, reaching 17.01\,m mean ATE on VBR, 6.17\,m on Oxford Spires, and 17.56\,m on KITTI---already outperforming calibrated DROID-W and uncalibrated SURE-Map on VBR as well as all uncalibrated geometric optimization baselines on Oxford Spires.

Finally, comparing Ours (Reconsolidation) against multi-stage geometric optimization and external loop-closure pipelines highlights the effectiveness of revising internal geometric context directly. Although calibrated optimization systems such as DROID-W and submap-alignment pipelines such as SURE-Map and CLoSeR leverage known intrinsics, explicit loop retrieval, or iterative bundle and pose-graph optimization, a single uncalibrated feed-forward reconsolidation pass achieves the lowest mean ATE across all evaluated methods on both VBR at 4.72\,m and KITTI at 9.42\,m. On Oxford Spires, Ours (Reconsolidation) reaches 3.92\,m mean ATE---the best among all uncalibrated methods---and achieves sub-meter trajectory error on eight of the ten campus sequences.

\begin{table}[H]
\centering
\caption{\textbf{Trajectory estimation on VBR.} Sequence and mean ATE (m). Within each group, bold/underline mark best/second best; across all methods, blue/pale blue mark best/second best (ties share color). LC denotes loop closure, and $^{*}$ indicates known intrinsics.}
\label{tab:vbr_comparison}
\small
\setlength{\tabcolsep}{4pt}
\setbox0=\hbox{\begin{tabular*}{\textwidth}{@{\extracolsep{\fill}}lrrrrrrrr@{}}
\toprule
& \multicolumn{7}{c}{Per-sequence ATE (m) $\downarrow$} & \\
\cmidrule(lr){2-8}
\hspace*{1em}Method & \multicolumn{1}{c}{\shortstack{Colosseo-0\\[-1pt]{\scriptsize\textcolor{gray}{8815f}}\\[-1pt]{\scriptsize\textcolor[rgb]{0,0,0.65}{1.45\,km}}}} & \multicolumn{1}{c}{\shortstack{Campus-0\\[-1pt]{\scriptsize\textcolor{gray}{12042f}}\\[-1pt]{\scriptsize\textcolor[rgb]{0,0,0.65}{2.73\,km}}}} & \multicolumn{1}{c}{\shortstack{Campus-1\\[-1pt]{\scriptsize\textcolor{gray}{11671f}}\\[-1pt]{\scriptsize\textcolor[rgb]{0,0,0.65}{2.95\,km}}}} & \multicolumn{1}{c}{\shortstack{Pincio-0\\[-1pt]{\scriptsize\textcolor{gray}{11142f}}\\[-1pt]{\scriptsize\textcolor[rgb]{0,0,0.65}{1.27\,km}}}} & \multicolumn{1}{c}{\shortstack{Spagna-0\\[-1pt]{\scriptsize\textcolor{gray}{14141f}}\\[-1pt]{\scriptsize\textcolor[rgb]{0,0,0.65}{1.56\,km}}}} & \multicolumn{1}{c}{\shortstack{Diag-0\\[-1pt]{\scriptsize\textcolor{gray}{10021f}}\\[-1pt]{\scriptsize\textcolor[rgb]{0,0,0.65}{1.02\,km}}}} & \multicolumn{1}{c}{\shortstack{Ciampino-1\\[-1pt]{\scriptsize\textcolor{gray}{18846f}}\\[-1pt]{\scriptsize\textcolor[rgb]{0,0,0.65}{5.20\,km}}}} & Avg. \\
\midrule
\multicolumn{9}{@{}l}{\textit{Offline}} \\
\hspace*{1em}VGGT & 39.01 & 113.09 & 85.20 & 76.94 & 61.31 & 34.44 & 198.49 & 86.93 \\
\hspace*{1em}VGGT-$\Omega$ & \cellcolor[HTML]{E8F1FA}\textbf{5.67} & \underline{111.39} & \textbf{77.30} & \textbf{22.87} & \textbf{21.41} & 26.97 & 193.90 & \textbf{65.65} \\
\hspace*{1em}DA3 & 24.41 & 125.95 & 93.47 & 73.31 & \underline{30.28} & 24.78 & 199.67 & 81.69 \\
\hspace*{1em}$\pi^{3}$ & 14.62 & 111.64 & 90.78 & 47.43 & 58.03 & \underline{19.96} & \underline{174.38} & 73.83 \\
\hspace*{1em}$\pi^{3}$X & \underline{5.69} & \textbf{100.63} & \underline{78.62} & \underline{26.31} & 58.43 & \textbf{19.24} & \textbf{174.18} & \underline{66.16} \\
\hspace*{1em}VGG-T$^{3}$ & 94.97 & 138.36 & 119.83 & 75.60 & 64.02 & 35.15 & 200.26 & 104.03 \\
\hspace*{1em}FastVGGT & 25.48 & 117.33 & 82.23 & 74.38 & 57.52 & 35.03 & 195.38 & 83.91 \\
\midrule
\multicolumn{9}{@{}l}{\textit{Geometric optimization}} \\
\hspace*{1em}VGGT-SLAM & 98.06 & 138.49 & 120.12 & 78.17 & 64.52 & 35.56 & 202.60 & 105.36 \\
\hspace*{1em}MASt3R-SLAM & 99.16 & 133.77 & 119.92 & 46.81 & 58.27 & 32.41 & 202.22 & 98.94 \\
\hspace*{1em}VGGT-Long & 44.13 & 132.68 & 115.62 & 64.30 & 59.05 & 33.66 & 179.42 & 89.84 \\
\hspace*{1em}DROID-SLAM$^{*}$ & 49.50 & 125.06 & 71.68 & 49.15 & 60.72 & 35.81 & 144.59 & 76.64 \\
\hspace*{1em}DPV-SLAM$^{*}$ & 74.14 & 106.73 & 57.16 & 48.04 & 51.67 & 35.75 & 87.32 & 65.83 \\
\hspace*{1em}DPVO$^{*}$ & 75.08 & 111.22 & 63.01 & 55.05 & 54.22 & 34.69 & 85.84 & 68.44 \\
\hspace*{1em}DROID-W$^{*}$ & \textbf{13.54} & \underline{30.94} & \underline{25.34} & \textbf{13.19} & \textbf{14.74} & \textbf{23.52} & \underline{60.58} & \textbf{25.98} \\
\hspace*{1em}SURE-Map & \underline{19.69} & \textbf{16.06} & \textbf{14.55} & \underline{32.19} & \underline{35.23} & \underline{24.41} & \textbf{57.86} & \underline{28.57} \\
\midrule
\multicolumn{9}{@{}l}{\textit{Streaming}} \\
\hspace*{1em}SLAM-Former (frontend) & 50.31 & 116.33 & 102.59 & 74.77 & 60.84 & 22.90 & 183.25 & 87.28 \\
\hspace*{1em}InfiniteVGGT & 89.94 & 136.82 & 116.67 & 77.72 & 63.35 & 33.94 & 202.12 & 102.94 \\
\hspace*{1em}OVGGT & 95.39 & 136.92 & 116.46 & 77.65 & 61.63 & 34.67 & 200.35 & 103.30 \\
\hspace*{1em}Stream3R-w & 92.51 & 138.38 & 116.38 & 77.43 & 64.57 & 35.64 & 202.14 & 103.87 \\
\hspace*{1em}CUT3R & 78.18 & 135.81 & 117.45 & 75.90 & 62.49 & 33.34 & 193.70 & 99.55 \\
\hspace*{1em}CUT3R w/ reset & 75.42 & 58.36 & 61.87 & 48.91 & 48.62 & 25.86 & 173.80 & 70.40 \\
\hspace*{1em}TTT3R & 99.12 & 130.01 & 108.55 & 75.77 & 59.51 & 35.51 & 187.04 & 99.36 \\
\hspace*{1em}TTT3R w/ reset & 73.48 & 56.42 & 52.21 & 37.03 & 34.91 & 31.11 & 173.36 & 65.50 \\
\hspace*{1em}LongStream & 83.00 & 120.29 & 106.92 & 71.39 & 60.81 & 30.58 & 177.51 & 92.93 \\
\hspace*{1em}LingBot-Map & \underline{19.32} & 18.49 & \underline{11.45} & 41.80 & 36.60 & \underline{22.54} & 67.64 & 31.12 \\
\hspace*{1em}HorizonStream & 40.98 & 28.17 & 24.54 & 26.89 & \underline{28.21} & 25.41 & \textbf{30.29} & \underline{29.21} \\
\hspace*{1em}ABot-Recon & 47.99 & \textbf{10.89} & 17.33 & \underline{21.29} & 29.30 & 28.02 & 54.38 & 29.89 \\
\cmidrule{1-9}
\hspace*{1em}\textbf{Ours (Online)} & \textbf{10.90} & \underline{12.17} & \textbf{9.25} & \textbf{14.78} & \cellcolor[HTML]{E8F1FA}\textbf{8.86} & \textbf{11.62} & \underline{51.49} & \textbf{17.01} \\
\midrule
\multicolumn{9}{@{}l}{\textit{Global refinement}} \\
\hspace*{1em}SLAM-Former (w/ backend) & 31.77 & 132.20 & 101.17 & 67.71 & 55.48 & 32.09 & 191.44 & 87.41 \\
\hspace*{1em}CLoSeR & 14.22 & \cellcolor[HTML]{E8F1FA}\underline{7.43} & \cellcolor[HTML]{E8F1FA}\underline{3.71} & 9.81 & \underline{13.04} & 12.94 & 16.94 & 11.16 \\
\hspace*{1em}Scal3R w/ LC & \underline{6.81} & 67.08 & 85.18 & 27.20 & 30.40 & 20.73 & 184.17 & 60.22 \\
\hspace*{1em}HorizonStream w/ LC & 24.77 & 27.55 & 8.81 & 16.48 & 21.35 & \cellcolor[HTML]{E8F1FA}\underline{7.36} & 18.12 & 17.78 \\
\hspace*{1em}ABot-Recon w/ LC & 16.70 & 8.06 & 4.30 & \cellcolor[HTML]{E8F1FA}\underline{5.15} & 17.91 & 9.68 & \cellcolor[HTML]{BFD9F2}\textbf{8.07} & \cellcolor[HTML]{E8F1FA}\underline{9.98} \\
\cmidrule{1-9}
\hspace*{1em}\textbf{Ours (Reconsolidation)} & \cellcolor[HTML]{BFD9F2}\textbf{2.06} & \cellcolor[HTML]{BFD9F2}\textbf{7.15} & \cellcolor[HTML]{BFD9F2}\textbf{3.46} & \cellcolor[HTML]{BFD9F2}\textbf{3.84} & \cellcolor[HTML]{BFD9F2}\textbf{2.00} & \cellcolor[HTML]{BFD9F2}\textbf{6.07} & \cellcolor[HTML]{E8F1FA}\underline{8.48} & \cellcolor[HTML]{BFD9F2}\textbf{4.72} \\
\bottomrule
\end{tabular*}}\typeout{VBR-TABLE-WIDTH=\the\wd0; REPORT-WIDTH=\the\textwidth}\ifdim\wd0>\textwidth\setbox0=\hbox{\resizebox{\textwidth}{!}{\box0}}\fi
\noindent\makebox[\textwidth][c]{\box0}
\end{table}

\clearpage

\begin{table}[H]
\centering
\caption{\textbf{Trajectory estimation on Oxford Spires.} Sequence and mean ATE (m), with frame counts shown in gray. Within each group, bold/underline mark best/second best; across all methods, blue/pale blue mark best/second best (ties share color). LC denotes loop closure, and $^{*}$ indicates known intrinsics. BL: Bodleian Library, CC: Christ Church, KC: Keble College, OQ: Observatory Quarter.}
\label{tab:oxford_full_ate}
\small
\setlength{\tabcolsep}{3.5pt}
\setbox0=\hbox{\begin{tabular*}{\textwidth}{@{\extracolsep{\fill}}lrrrrrrrrrrr@{}}
\toprule
& \multicolumn{10}{c}{Per-sequence ATE (m) $\downarrow$} & \\
\cmidrule(lr){2-11}
\hspace*{1em}Method & \multicolumn{1}{c}{\shortstack{KC-02\\[-1pt]{\scriptsize\textcolor{gray}{5969f}}}} & \multicolumn{1}{c}{\shortstack{KC-03\\[-1pt]{\scriptsize\textcolor{gray}{5682f}}}} & \multicolumn{1}{c}{\shortstack{KC-04\\[-1pt]{\scriptsize\textcolor{gray}{13518f}}}} & \multicolumn{1}{c}{\shortstack{KC-05\\[-1pt]{\scriptsize\textcolor{gray}{11544f}}}} & \multicolumn{1}{c}{\shortstack{OQ-01\\[-1pt]{\scriptsize\textcolor{gray}{5746f}}}} & \multicolumn{1}{c}{\shortstack{OQ-02\\[-1pt]{\scriptsize\textcolor{gray}{5469f}}}} & \multicolumn{1}{c}{\shortstack{CC-02\\[-1pt]{\scriptsize\textcolor{gray}{11775f}}}} & \multicolumn{1}{c}{\shortstack{CC-03\\[-1pt]{\scriptsize\textcolor{gray}{6204f}}}} & \multicolumn{1}{c}{\shortstack{CC-05\\[-1pt]{\scriptsize\textcolor{gray}{3961f}}}} & \multicolumn{1}{c}{\shortstack{BL-02\\[-1pt]{\scriptsize\textcolor{gray}{4966f}}}} & Avg. \\
\midrule
\multicolumn{12}{@{}l}{\textit{Offline}} \\
\hspace*{1em}VGGT & 18.26 & 32.08 & 18.49 & 28.62 & 11.07 & 9.34 & 32.20 & 8.86 & 42.27 & 71.72 & 27.29 \\
\hspace*{1em}VGGT-$\Omega$ & \textbf{2.19} & \textbf{2.63} & \cellcolor[HTML]{E8F1FA}\textbf{1.78} & \textbf{4.66} & \textbf{1.62} & \textbf{1.03} & \textbf{14.07} & \textbf{0.69} & \underline{35.37} & \underline{37.04} & \textbf{10.11} \\
\hspace*{1em}DA3 & 3.12 & 10.27 & 6.07 & 21.28 & 8.25 & 8.68 & 40.55 & \underline{1.49} & 36.02 & 57.57 & 19.33 \\
\hspace*{1em}$\pi^{3}$ & 5.81 & 13.59 & 4.76 & \underline{20.30} & 4.61 & 3.81 & 39.14 & 2.70 & \textbf{32.32} & 68.98 & 19.60 \\
\hspace*{1em}$\pi^{3}$X & \underline{2.98} & \underline{6.49} & \underline{2.18} & 24.61 & \underline{2.70} & \underline{2.27} & \underline{24.90} & 3.15 & 41.73 & \textbf{21.30} & \underline{13.23} \\
\hspace*{1em}VGG-T$^{3}$ & 28.87 & 32.34 & 42.49 & 41.46 & 27.52 & 26.70 & 43.42 & 16.96 & 40.88 & 73.54 & 37.42 \\
\hspace*{1em}FastVGGT & 18.42 & 32.22 & 14.52 & 24.47 & 10.53 & 7.18 & 34.97 & 7.98 & 41.64 & 75.42 & 26.73 \\
\midrule
\multicolumn{12}{@{}l}{\textit{Geometric optimization}} \\
\hspace*{1em}VGGT-SLAM & 25.86 & \underline{2.27} & 15.05 & 41.07 & 5.61 & 3.82 & 43.12 & 17.62 & 42.59 & 82.34 & 27.94 \\
\hspace*{1em}MASt3R-SLAM & 5.06 & 31.40 & \textbf{2.39} & 37.37 & 2.25 & \underline{1.67} & 16.84 & 0.70 & 37.73 & 34.11 & 16.95 \\
\hspace*{1em}VGGT-Long & \underline{1.76} & 14.74 & 32.51 & \underline{4.60} & 3.69 & 5.66 & 11.59 & 3.99 & \underline{13.51} & 16.16 & 10.82 \\
\hspace*{1em}DROID-SLAM$^{*}$ & 6.71 & 31.03 & \underline{2.73} & 16.42 & \cellcolor[HTML]{BFD9F2}\textbf{0.17} & \cellcolor[HTML]{BFD9F2}\textbf{0.21} & 14.39 & \underline{0.47} & 17.18 & \cellcolor[HTML]{BFD9F2}\textbf{0.61} & 8.99 \\
\hspace*{1em}DPV-SLAM$^{*}$ & 8.77 & 4.27 & 15.88 & 27.94 & \cellcolor[HTML]{E8F1FA}\underline{0.32} & 11.75 & \cellcolor[HTML]{E8F1FA}\textbf{1.02} & 0.93 & 39.73 & 3.17 & 11.38 \\
\hspace*{1em}DPVO$^{*}$ & 7.30 & 25.84 & 19.12 & 26.11 & 0.80 & 4.84 & \underline{1.13} & \cellcolor[HTML]{E8F1FA}\textbf{0.43} & 42.61 & \underline{3.12} & 13.13 \\
\hspace*{1em}DROID-W$^{*}$ & \cellcolor[HTML]{E8F1FA}\textbf{0.82} & \cellcolor[HTML]{E8F1FA}\textbf{1.30} & 4.49 & \textbf{2.67} & 2.21 & 2.87 & 2.71 & 0.72 & \cellcolor[HTML]{BFD9F2}\textbf{1.92} & 6.73 & \cellcolor[HTML]{BFD9F2}\textbf{2.64} \\
\hspace*{1em}SURE-Map & 2.05 & 3.93 & 3.47 & 9.14 & 3.54 & 3.71 & 13.60 & 3.88 & 31.92 & 8.92 & \underline{8.42} \\
\midrule
\multicolumn{12}{@{}l}{\textit{Streaming}} \\
\hspace*{1em}SLAM-Former (frontend) & 28.20 & 25.20 & 26.12 & 28.40 & 16.39 & 15.39 & 34.29 & 12.21 & 32.82 & 69.06 & 28.81 \\
\hspace*{1em}InfiniteVGGT & 26.51 & 28.87 & 39.27 & 36.14 & 28.67 & 27.39 & 39.20 & 16.99 & 38.61 & 77.72 & 35.94 \\
\hspace*{1em}OVGGT & 31.09 & 30.85 & 39.76 & 31.85 & 28.38 & 27.79 & 38.97 & 17.31 & 40.43 & 77.36 & 36.38 \\
\hspace*{1em}Stream3R-w & 31.25 & 31.44 & 41.47 & 36.86 & 28.58 & 26.46 & 40.50 & 17.18 & 41.85 & 79.01 & 37.46 \\
\hspace*{1em}CUT3R & 26.76 & 27.01 & 40.04 & 36.41 & 27.40 & 25.37 & 39.46 & 16.27 & 41.97 & 76.45 & 35.71 \\
\hspace*{1em}CUT3R w/ reset & 20.11 & 13.88 & 19.55 & 32.17 & 18.19 & 19.23 & 28.39 & 6.09 & 31.64 & 29.62 & 21.89 \\
\hspace*{1em}TTT3R & 23.18 & 21.74 & 33.79 & 39.50 & 29.01 & 28.77 & 35.89 & 12.47 & 33.74 & 49.98 & 30.81 \\
\hspace*{1em}TTT3R w/ reset & 14.75 & 8.65 & 12.79 & 26.01 & 11.91 & 12.67 & 16.36 & 2.38 & \underline{22.71} & 22.62 & 15.08 \\
\hspace*{1em}LongStream & 20.02 & 17.69 & 30.00 & 33.65 & 19.73 & 18.95 & 35.60 & 11.01 & 34.43 & 42.69 & 26.38 \\
\hspace*{1em}LingBot-Map & \underline{3.40} & \underline{3.25} & \underline{2.99} & 9.33 & \underline{2.80} & \underline{2.49} & 14.15 & 2.95 & 31.81 & \underline{9.35} & 8.25 \\
\hspace*{1em}HorizonStream & 13.59 & 8.70 & 20.29 & 30.82 & 9.66 & 9.27 & 31.05 & 6.11 & 26.12 & 28.25 & 18.39 \\
\hspace*{1em}ABot-Recon & 5.28 & 4.26 & 10.30 & \underline{8.47} & 6.01 & 5.28 & \textbf{11.60} & \underline{1.98} & \textbf{9.30} & 11.34 & \underline{7.38} \\
\cmidrule{1-12}
\hspace*{1em}\textbf{Ours (Online)} & \textbf{1.43} & \textbf{1.60} & \textbf{1.91} & \cellcolor[HTML]{E8F1FA}\textbf{1.97} & \textbf{1.66} & \textbf{1.62} & \underline{13.41} & \textbf{0.51} & 28.47 & \textbf{9.11} & \textbf{6.17} \\
\midrule
\multicolumn{12}{@{}l}{\textit{Global refinement}} \\
\hspace*{1em}SLAM-Former (w/ backend) & 6.25 & 30.70 & 17.00 & 39.93 & 11.86 & 12.00 & 39.38 & 3.28 & 38.07 & 67.17 & 26.56 \\
\hspace*{1em}CLoSeR & 2.60 & \underline{2.32} & 5.58 & \underline{4.58} & 4.32 & 4.04 & \underline{5.57} & \underline{1.18} & \cellcolor[HTML]{E8F1FA}\textbf{6.82} & \underline{7.46} & \underline{4.45} \\
\hspace*{1em}Scal3R w/ LC & 9.72 & 4.50 & \underline{2.09} & 12.98 & \underline{0.95} & \underline{0.51} & 22.81 & 9.07 & 36.75 & 11.43 & 11.08 \\
\hspace*{1em}HorizonStream w/ LC & 6.44 & 8.69 & 19.16 & 21.47 & 5.75 & 1.52 & 30.01 & 5.76 & 26.14 & 28.25 & 15.32 \\
\hspace*{1em}ABot-Recon w/ LC & \underline{2.46} & 4.26 & 5.83 & 4.68 & 3.34 & 2.73 & 7.87 & 1.92 & \underline{9.30} & 11.34 & 5.37 \\
\cmidrule{1-12}
\hspace*{1em}\textbf{Ours (Reconsolidation)} & \cellcolor[HTML]{BFD9F2}\textbf{0.44} & \cellcolor[HTML]{BFD9F2}\textbf{0.42} & \cellcolor[HTML]{BFD9F2}\textbf{0.55} & \cellcolor[HTML]{BFD9F2}\textbf{0.69} & \cellcolor[HTML]{E8F1FA}\textbf{0.32} & \cellcolor[HTML]{E8F1FA}\textbf{0.40} & \cellcolor[HTML]{BFD9F2}\textbf{1.00} & \cellcolor[HTML]{BFD9F2}\textbf{0.24} & 32.29 & \cellcolor[HTML]{E8F1FA}\textbf{2.88} & \cellcolor[HTML]{E8F1FA}\textbf{3.92} \\
\bottomrule
\end{tabular*}}\typeout{OXFORD-TABLE-WIDTH=\the\wd0; REPORT-WIDTH=\the\textwidth}\ifdim\wd0>\textwidth\setbox0=\hbox{\resizebox{\textwidth}{!}{\box0}}\fi
\noindent\makebox[\textwidth][c]{\box0}
\end{table}

\clearpage

\begin{table}[H]
\centering
\caption{\textbf{Trajectory estimation on KITTI.} Sequence and mean ATE (m). Within each group, bold/underline mark best/second best; across all methods, blue/pale blue mark best/second best (ties share color). LC denotes loop closure, and $^{*}$ indicates known intrinsics.}
\label{tab:kitti_comparison}
\small
\setlength{\tabcolsep}{1.8pt}
\setbox0=\hbox{\begin{tabular*}{\textwidth}{@{\extracolsep{\fill}}lrrrrrrrrrrrr@{}}
\toprule
& \multicolumn{11}{c}{Per-sequence ATE (m) $\downarrow$} & \\
\cmidrule(lr){2-12}
\hspace*{1em}Method & \multicolumn{1}{c}{\shortstack{00\\[-1pt]{\scriptsize\textcolor{gray}{4541f}}\\[-1pt]{\scriptsize\textcolor[rgb]{0,0,0.65}{3.70\,km}}}} & \multicolumn{1}{c}{\shortstack{01\\[-1pt]{\scriptsize\textcolor{gray}{1101f}}\\[-1pt]{\scriptsize\textcolor[rgb]{0,0,0.65}{2.50\,km}}}} & \multicolumn{1}{c}{\shortstack{02\\[-1pt]{\scriptsize\textcolor{gray}{4661f}}\\[-1pt]{\scriptsize\textcolor[rgb]{0,0,0.65}{5.10\,km}}}} & \multicolumn{1}{c}{\shortstack{03\\[-1pt]{\scriptsize\textcolor{gray}{801f}}\\[-1pt]{\scriptsize\textcolor[rgb]{0,0,0.65}{0.60\,km}}}} & \multicolumn{1}{c}{\shortstack{04\\[-1pt]{\scriptsize\textcolor{gray}{271f}}\\[-1pt]{\scriptsize\textcolor[rgb]{0,0,0.65}{0.40\,km}}}} & \multicolumn{1}{c}{\shortstack{05\\[-1pt]{\scriptsize\textcolor{gray}{2761f}}\\[-1pt]{\scriptsize\textcolor[rgb]{0,0,0.65}{2.20\,km}}}} & \multicolumn{1}{c}{\shortstack{06\\[-1pt]{\scriptsize\textcolor{gray}{1101f}}\\[-1pt]{\scriptsize\textcolor[rgb]{0,0,0.65}{1.20\,km}}}} & \multicolumn{1}{c}{\shortstack{07\\[-1pt]{\scriptsize\textcolor{gray}{1101f}}\\[-1pt]{\scriptsize\textcolor[rgb]{0,0,0.65}{0.70\,km}}}} & \multicolumn{1}{c}{\shortstack{08\\[-1pt]{\scriptsize\textcolor{gray}{4071f}}\\[-1pt]{\scriptsize\textcolor[rgb]{0,0,0.65}{3.20\,km}}}} & \multicolumn{1}{c}{\shortstack{09\\[-1pt]{\scriptsize\textcolor{gray}{1591f}}\\[-1pt]{\scriptsize\textcolor[rgb]{0,0,0.65}{1.70\,km}}}} & \multicolumn{1}{c}{\shortstack{10\\[-1pt]{\scriptsize\textcolor{gray}{1201f}}\\[-1pt]{\scriptsize\textcolor[rgb]{0,0,0.65}{0.90\,km}}}} & Avg. \\
\midrule
\multicolumn{13}{@{}l}{\textit{Offline}} \\
\hspace*{1em}VGGT & 187.55 & 604.34 & 297.93 & 22.74 & 6.90 & 140.79 & 32.18 & 51.82 & 248.30 & 205.46 & 186.73 & 180.43 \\
\hspace*{1em}VGGT-$\Omega$ & \textbf{170.72} & \textbf{310.95} & \textbf{286.99} & 10.06 & \textbf{1.35} & \textbf{55.97} & \underline{10.00} & \textbf{10.72} & \underline{210.02} & \textbf{104.21} & \textbf{173.12} & \textbf{122.19} \\
\hspace*{1em}DA3 & 181.61 & 610.23 & 300.62 & 48.49 & 12.49 & 154.91 & 53.26 & 33.30 & 256.72 & \underline{191.21} & \underline{175.70} & 183.50 \\
\hspace*{1em}$\pi^{3}$ & 178.24 & 511.47 & 301.96 & \underline{7.45} & 2.25 & 148.55 & 29.36 & 47.74 & 220.68 & 208.11 & 177.80 & 166.69 \\
\hspace*{1em}$\pi^{3}$X & \underline{174.13} & \underline{503.62} & \underline{295.08} & \textbf{3.17} & \underline{1.70} & \underline{118.38} & \textbf{5.12} & \underline{22.09} & \textbf{200.64} & 207.49 & 176.78 & \underline{155.29} \\
\hspace*{1em}VGG-T$^{3}$ & 191.88 & 701.05 & 303.77 & 110.63 & 112.99 & 158.82 & 134.42 & 85.41 & 251.33 & 201.59 & 208.35 & 223.66 \\
\hspace*{1em}FastVGGT & 188.42 & 543.88 & 302.07 & 24.92 & 9.92 & 133.82 & 45.93 & 50.46 & 252.72 & 206.02 & 186.31 & 176.77 \\
\midrule
\multicolumn{13}{@{}l}{\textit{Geometric optimization}} \\
\hspace*{1em}VGGT-SLAM & 193.30 & 507.63 & 269.06 & 168.98 & 3.38 & 159.92 & 40.80 & 23.25 & 264.27 & 219.41 & 209.65 & 187.24 \\
\hspace*{1em}MASt3R-SLAM & 145.43 & 556.15 & 279.09 & 6.00 & 97.80 & 144.41 & 130.59 & 77.43 & 263.65 & 162.09 & 151.77 & 183.13 \\
\hspace*{1em}VGGT-Long & \underline{10.83} & 66.50 & 55.69 & 5.58 & 3.95 & \underline{10.13} & 6.95 & \textbf{2.83} & 57.55 & 31.22 & 22.15 & \underline{24.85} \\
\hspace*{1em}DROID-SLAM$^{*}$ & 92.66 & 87.09 & 140.40 & 2.39 & \underline{0.97} & 54.13 & 64.87 & 21.11 & 63.39 & 73.22 & 15.88 & 56.01 \\
\hspace*{1em}DPV-SLAM$^{*}$ & 113.28 & \cellcolor[HTML]{BFD9F2}\textbf{11.21} & 117.29 & \underline{2.28} & \textbf{0.94} & 57.32 & 55.96 & 18.70 & 109.26 & 74.80 & 13.26 & 52.21 \\
\hspace*{1em}DPVO$^{*}$ & 113.23 & \cellcolor[HTML]{E8F1FA}\underline{12.49} & 117.38 & \cellcolor[HTML]{E8F1FA}\textbf{2.26} & \textbf{0.94} & 57.10 & 56.06 & 18.99 & 109.41 & 74.66 & 13.28 & 52.35 \\
\hspace*{1em}DROID-W$^{*}$ & \textbf{8.24} & 212.74 & \cellcolor[HTML]{BFD9F2}\textbf{14.88} & 2.82 & 4.64 & \textbf{8.40} & \underline{4.69} & \underline{3.97} & \underline{17.08} & \textbf{15.00} & \cellcolor[HTML]{BFD9F2}\textbf{5.50} & 27.09 \\
\hspace*{1em}SURE-Map & 20.80 & 39.53 & \underline{53.47} & \underline{2.28} & 1.34 & 18.44 & \textbf{4.00} & 3.99 & \textbf{14.38} & \underline{21.14} & \underline{10.19} & \textbf{17.23} \\
\midrule
\multicolumn{13}{@{}l}{\textit{Streaming}} \\
\hspace*{1em}SLAM-Former (frontend) & 190.12 & 567.72 & 283.74 & 130.02 & 100.80 & 154.24 & 94.15 & 55.17 & 220.71 & 202.64 & 194.29 & 199.42 \\
\hspace*{1em}InfiniteVGGT & 186.41 & 623.20 & 289.31 & 166.86 & 68.12 & 143.86 & 117.34 & 85.29 & 221.41 & 215.35 & 156.84 & 206.73 \\
\hspace*{1em}OVGGT & 184.76 & 717.61 & 304.78 & 157.05 & 65.34 & 156.04 & 107.28 & 86.29 & 247.45 & 219.27 & 178.49 & 220.40 \\
\hspace*{1em}Stream3R-w & 188.66 & 695.02 & 301.28 & 161.68 & 98.99 & 159.85 & 122.82 & 87.19 & 264.61 & 219.90 & 194.32 & 226.76 \\
\hspace*{1em}CUT3R & 189.88 & 648.13 & 276.51 & 148.03 & 22.61 & 151.43 & 132.07 & 79.85 & 235.75 & 176.04 & 188.03 & 204.39 \\
\hspace*{1em}CUT3R w/ reset & 191.08 & 101.71 & 266.74 & 20.74 & 7.66 & 106.04 & 69.25 & 23.51 & 160.85 & 71.78 & 37.46 & 96.07 \\
\hspace*{1em}TTT3R & 177.72 & 539.47 & 272.82 & 100.90 & 10.95 & 147.40 & 131.13 & 65.99 & 230.32 & 179.40 & 119.81 & 179.63 \\
\hspace*{1em}TTT3R w/ reset & 113.17 & 100.15 & 224.31 & 15.90 & 4.50 & 31.01 & 41.15 & 11.27 & 93.45 & 87.63 & 33.88 & 68.76 \\
\hspace*{1em}LongStream & 90.02 & 63.45 & 236.44 & 5.03 & 2.57 & 80.62 & 12.77 & 13.84 & 89.59 & 95.06 & 25.65 & 65.00 \\
\hspace*{1em}LingBot-Map & 23.41 & 88.76 & 71.51 & \textbf{2.80} & \underline{0.88} & 17.39 & \underline{4.82} & 5.11 & \cellcolor[HTML]{E8F1FA}\textbf{14.04} & 21.64 & 14.45 & 24.07 \\
\hspace*{1em}HorizonStream & 29.52 & \underline{24.09} & 97.40 & 5.55 & \cellcolor[HTML]{E8F1FA}\textbf{0.72} & \underline{13.90} & 5.49 & 6.10 & 22.59 & 29.77 & \underline{12.39} & 22.50 \\
\hspace*{1em}ABot-Recon & \underline{22.43} & \textbf{21.59} & \underline{70.06} & \underline{4.36} & 2.91 & \textbf{13.38} & 12.05 & \textbf{1.98} & 22.40 & \underline{14.12} & 16.27 & \underline{18.32} \\
\cmidrule{1-13}
\hspace*{1em}\textbf{Ours (Online)} & \textbf{20.13} & 64.35 & \textbf{46.49} & 4.54 & 1.09 & 15.46 & \cellcolor[HTML]{E8F1FA}\textbf{2.87} & \underline{3.92} & \underline{14.57} & \textbf{12.99} & \textbf{6.70} & \textbf{17.56} \\
\midrule
\multicolumn{13}{@{}l}{\textit{Global refinement}} \\
\hspace*{1em}SLAM-Former (w/ backend) & 189.14 & 687.55 & 297.39 & 133.26 & 58.92 & 153.69 & 100.98 & 69.68 & 247.02 & 180.58 & 209.66 & 211.62 \\
\hspace*{1em}CLoSeR & 6.55 & 41.78 & \underline{28.53} & 4.86 & 1.82 & 6.55 & 4.76 & 4.87 & 26.51 & \cellcolor[HTML]{BFD9F2}\textbf{5.66} & \underline{8.79} & \cellcolor[HTML]{E8F1FA}\underline{12.79} \\
\hspace*{1em}Scal3R w/ LC & \cellcolor[HTML]{BFD9F2}\textbf{4.17} & 50.47 & 37.76 & 5.21 & 1.05 & \cellcolor[HTML]{BFD9F2}\textbf{3.84} & \underline{3.36} & \cellcolor[HTML]{E8F1FA}\underline{1.37} & 25.29 & 12.43 & \cellcolor[HTML]{E8F1FA}\textbf{5.94} & 13.72 \\
\hspace*{1em}HorizonStream w/ LC & 15.37 & \underline{24.09} & 71.07 & 5.55 & \cellcolor[HTML]{E8F1FA}\underline{0.72} & 7.83 & 7.63 & 3.14 & 22.59 & 27.45 & 12.38 & 17.98 \\
\hspace*{1em}ABot-Recon w/ LC & 11.68 & \textbf{21.59} & 39.83 & \underline{4.36} & 2.91 & 5.89 & 8.65 & 1.65 & \underline{22.40} & 13.98 & 16.27 & 13.56 \\
\cmidrule{1-13}
\hspace*{1em}\textbf{Ours (Reconsolidation)} & \cellcolor[HTML]{E8F1FA}\underline{5.64} & 39.77 & \cellcolor[HTML]{E8F1FA}\textbf{22.97} & \cellcolor[HTML]{BFD9F2}\textbf{1.16} & \cellcolor[HTML]{BFD9F2}\textbf{0.58} & \cellcolor[HTML]{E8F1FA}\underline{5.49} & \cellcolor[HTML]{BFD9F2}\textbf{1.19} & \cellcolor[HTML]{BFD9F2}\textbf{1.30} & \cellcolor[HTML]{BFD9F2}\textbf{9.21} & \cellcolor[HTML]{E8F1FA}\underline{6.97} & 9.36 & \cellcolor[HTML]{BFD9F2}\textbf{9.42} \\
\bottomrule
\end{tabular*}}\typeout{KITTI-TABLE-WIDTH=\the\wd0; REPORT-WIDTH=\the\textwidth}\ifdim\wd0>\textwidth\setbox0=\hbox{\resizebox{\textwidth}{!}{\box0}}\fi
\noindent\makebox[\textwidth][c]{\box0}
\end{table}

\clearpage
\begin{figure}[p]
\centering
\includegraphics[width=\linewidth]{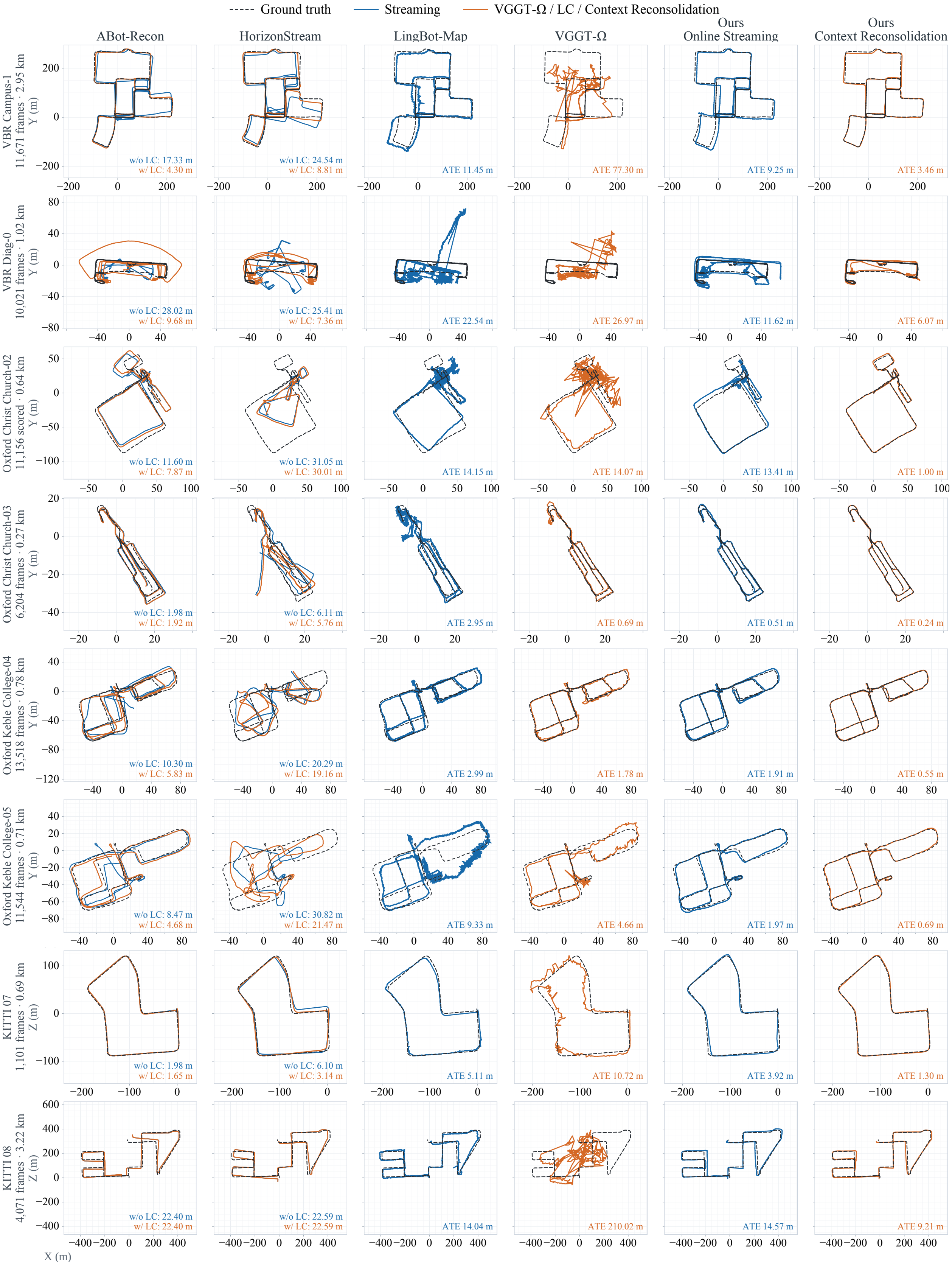}
\captionsetup{skip=3pt}
\caption{\textbf{Additional long-trajectory comparisons.} Independently Sim(3)-aligned to GT; shared XY (VBR/Oxford) or XZ (KITTI) axes per row. Axes and ATE are in meters; LC denotes loop closure.}
\label{fig:supp_trajectory_comparison}
\end{figure}

\clearpage
\section{Additional Qualitative Results}
\label{sec:supp_qualitative}

\begin{figure}[H]
\centering
\includegraphics[width=\textwidth]{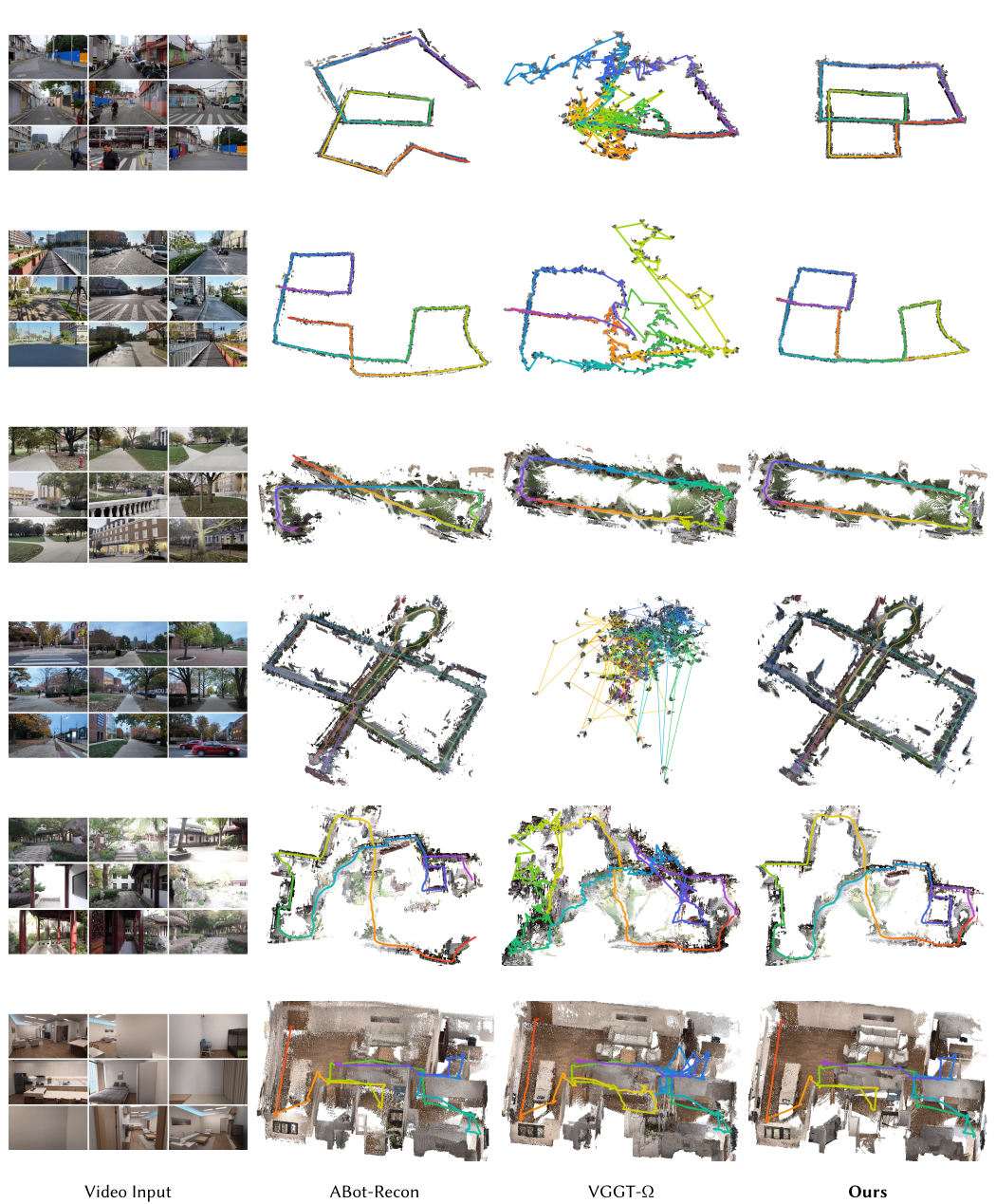}
\caption{\textbf{Qualitative 3D reconstruction comparisons (continued).} Six additional sequences, with sampled input frames and reconstructions from ABot-Recon, VGGT-$\Omega$, and \method in the same column order as \cref{fig:more_3d_reconstruction}.}
\label{fig:more_3d_reconstruction_extra}
\end{figure}

\end{document}